\documentclass[lettersize,journal]{IEEEtran}
\usepackage{amsmath,amsfonts}
\usepackage{algorithmic}
\usepackage{algorithm}
\usepackage{array}
\usepackage[caption=false,font=normalsize,labelfont=sf,textfont=sf]{subfig}
\usepackage{textcomp}
\usepackage{stfloats}
\usepackage{url}
\usepackage{verbatim}
\usepackage{graphicx}
\usepackage{xcolor}
\usepackage{cite}
\begin{document}

\title{RAZOR: Refining Accuracy\\ by Zeroing Out Redundancies}

\author{
    Daniel Riccio,~\IEEEmembership{Member,~IEEE,}
    Genoveffa Tortora,~\IEEEmembership{Senior Member,~IEEE,}
    Mara Sangiovanni,~\IEEEmembership{Member,~IEEE}
    \thanks{Daniel Riccio is with the University of Naples Federico II, Naples, Italy (email: daniel.riccio@unina.it).}
    \thanks{Genoveffa Tortora is with the University of Salerno, Salerno, Italy (email: tortora@unisa.it).}
    \thanks{Mara Sangiovanni is with the University of Naples Federico II, Naples, Italy (email: mara.sangiovanni@unina.it).}
}

\markboth{Journal of \LaTeX\ Class Files,~Vol.~, No.~, September~2024}%
{Shell \MakeLowercase{\textit{et al.}}: RAZOR: Refining Accuracy by Zeroing Out Redundancies}

\IEEEpubid{0\$0~\copyright~}
This work has been submitted to the IEEE for possible publication. Copyright may be transferred without notice, after which this version may no longer be accessible.

\maketitle

\begin{abstract}
In many application domains, the proliferation of sensors and devices is generating vast volumes of data, imposing significant pressure on existing data analysis and data mining techniques. Nevertheless, an increase in data volume does not inherently imply an increase in informational content, as a substantial portion may be redundant or represent noise. This challenge is particularly evident in the deep learning domain, where the utility of additional data is contingent on its informativeness. In the absence of such, larger datasets merely exacerbate the computational cost and complexity of the learning process. To address these challenges, we propose RAZOR, a novel instance selection technique designed to extract a significantly smaller yet sufficiently informative subset from a larger set of instances without compromising the learning process. RAZOR has been specifically engineered to be robust, efficient, and scalable, making it suitable for large-scale datasets. Unlike many techniques in the literature, RAZOR is capable of operating in both supervised and unsupervised settings. Experimental results demonstrate that RAZOR outperforms recent state-of-the-art techniques in terms of both effectiveness and efficiency.
\end{abstract}


\section{Introduction}
\IEEEPARstart{I}{n} data mining, both supervised and unsupervised learning are essential. Although deep learning models have made significant progress, allowing for the autonomous extraction of features and classifiers from raw data, managing the vast and unstructured datasets commonly encountered today still presents substantial challenges. These datasets, often immense in size, rapidly accumulating, and inherently complex, can overwhelm traditional analytical methods, making them inefficient for processing such large-scale data.

As a response to these challenges, data preprocessing—particularly data reduction—has gained increasing importance. Data reduction focuses on eliminating irrelevant, redundant, or noisy information, setting it apart from broader data preparation techniques. This step is essential for improving the efficiency of models used for tasks such as classification, segmentation, or regression. By refining the dataset in this way, the computational load is significantly reduced, enabling learning algorithms to perform more effectively, especially when handling large-scale data.

The four most commonly employed techniques for reducing the size of training datasets are dataset pruning, few-shot learning and data selection. Dataset pruning involves reducing the size of the training data while maintaining high model performance~\cite{raju2021accelerating}. This method differs from few-shot learning, which focuses on achieving strong performance using minimal amounts of training data. Data selection methods, often applied in the context of continual and active learning~\cite{sener2017active}, involve selecting a subset of training samples based on specific criteria, such as compactness, diversity, forgetfulness, or gradient norm. Dataset distillation and condensation aim to synthesize smaller, information-rich datasets that serve as efficient alternatives to larger datasets~\cite{yu2023dataset}. Lastly, the influence function is a statistical machine learning approach that evaluates how the removal of specific training examples impacts the model's generalization ability~\cite{yang2022dataset}.

Among data reduction techniques, data selection offers a notable advantage in striking a balance between efficiency and effectiveness. Unlike dataset pruning, which merely reduces the size of the training data while maintaining performance, data selection specifically identifies and targets the most informative and representative samples. Furthermore, in contrast to few-shot learning, which seeks to achieve optimal performance using minimal data, data selection adopts a more pragmatic approach by optimizing the existing dataset, rather than facing the inherent challenges of learning from extremely limited data.
Within data selection methods, Instance Selection (IS) stands out as particularly advantageous due to its ability to select individual instances based on criteria such as compactness, diversity, and relevance to the target task. This approach proves especially valuable in scenarios involving large, unstructured datasets, where computational resources are limited and maintaining high data quality is crucial. By concentrating on the most relevant data, instance selection facilitates the development of models with superior generalization capabilities, ultimately leading to enhanced performance in predictive tasks.
A significant limitation of many instance selection methods is their dependence on supervised selection, which requires access to labeled data. This reliance can be impractical in situations where labeled instances are scarce or difficult to obtain. As an alternative, unsupervised data pruning challenges the traditional scaling law in machine learning, which posits that improvements in performance are typically achieved by increasing model parameters, dataset size, or computational power. Conventional methods adhering to this law often yield only marginal error reductions, despite substantial investments in both data and computational resources.

In contrast, unsupervised data pruning offers a more sustainable and efficient approach, enabling effective compression of training data without compromising model performance. This technique is particularly advantageous in contexts where labeled data is either unavailable or prohibitively costly to obtain. By promoting a more intelligent and resource-efficient strategy for managing training data, unsupervised data pruning represents an evolution in the field, addressing a range of real-world applications where labeled data is limited, and traditional data scaling methods are insufficient.

\section{State of the Art}
The literature on data reduction in data mining and machine learning is characterized by a rich diversity of techniques, each aimed at enhancing and refining datasets to optimize learning outcomes. These techniques can be categorized into three primary classes: (a) those that focus on sample relevance, (b) those that modify sample labels, and (c) those that reduce the number of samples. These categories represent distinct yet complementary approaches to data management and optimization, contributing to both improved computational efficiency and enhanced model accuracy in machine learning.

The first class encompasses methods designed to refine classifiers for processing complex data structures effectively. This includes techniques such as boosting and margin maximization, exemplified by methods like AdaBoost and Support Vector Machines (SVMs), which are designed to improve classifier performance~\cite{freund1997decision,cortes1995support}. Boosting methods iteratively refine classifiers by adjusting instance weights, placing greater emphasis on difficult-to-classify instances~\cite{schapire1999brief}. While this technique can significantly enhance performance, it may also lead to overfitting and increased computational demands. On the other hand, margin maximization, employed in SVMs, seeks to maximize the margin between classes to improve generalization. However, this approach is computationally intensive and requires careful parameter tuning to avoid suboptimal outcomes~\cite{burges1998tutorial}.
Another strategy within this class is ranking-based approaches, which prioritize instances based on their relevance or importance, with the goal of emphasizing key data points that could enhance model performance~\cite{joachims2002optimizing}. However, the subjective nature of ranking criteria can introduce biases, potentially affecting generalization. Therefore, aligning ranking criteria with the specific learning task and the characteristics of the dataset is crucial to mitigate these challenges~\cite{guyon2003introduction}.

A more innovative segment of this class includes decremental optimization, evolutionary strategies, agent-based models, and neural network-based methods, as well as approaches that leverage SVMs~\cite{holland1992adaptation,koza1992genetic,lecun1998gradient,hinton2006reducing}. While each method offers distinct advantages, they also present specific challenges such as high computational complexity, sensitivity to parameter tuning, susceptibility to overfitting, and the need for sophisticated setups~\cite{lecun1998gradient,hinton2006reducing}. Additionally, the influence function is a tool that evaluates the effect of removing individual data points on model performance. It enables the identification and elimination of instances that negatively impact model accuracy, offering an effective way to refine datasets~\cite{koh2017understanding}.

The second class of data reduction techniques focuses on editing methods, which refine datasets by identifying and correcting erroneously labeled instances. These techniques enhance data quality by ensuring that each instance is accurately categorized, a crucial factor in supervised learning contexts where label accuracy directly influences model performance. The primary benefit of editing methods lies in their ability to create a more reliable and coherent dataset, thereby significantly improving both model accuracy and generalizability~\cite{zhu2004class,garcia2012survey}. However, these methods are often computationally demanding, requiring detailed analysis and iterative reviews, particularly in the case of large datasets. Additionally, a key risk associated with editing is the potential for over-editing, which may result in the removal of valuable information. This is particularly problematic when outliers or unique instances, which may be critical for model accuracy, are erroneously discarded~\cite{tomek1976two,zhang2003editing}.

The third class is broader in scope and encompasses several approaches, including dataset distillation and condensation, dataset pruning, and data selection.
Condensation techniques aim to reduce dataset size while preserving its core classification characteristics, making them especially useful for managing large datasets~\cite{nguyen2021sample}. These methods streamline the data analysis process by minimizing both storage and computational requirements, which is particularly advantageous when handling large volumes of data. However, a notable drawback of condensation techniques is their tendency to overlook subtle but important patterns within the data. In the process of reducing dataset size, these methods may inadvertently discard instances that appear redundant but contain valuable information for classification. This loss can diminish the richness and diversity of the dataset, potentially compromising the robustness and accuracy of the resulting classification models.
Merging techniques, in contrast, aim to reduce redundancy by combining similar instances within the dataset~\cite{zhang2003editing}. By aggregating sufficiently similar instances, these methods condense the dataset, making it more manageable while reducing both storage and computational burdens during the training of machine learning models. This approach is particularly effective for voluminous datasets that contain repetitive or closely related instances. However, the primary risk associated with merging techniques is the potential loss of distinctive features. Combining similar instances may obscure or eliminate unique characteristics that are critical for specific classification or prediction tasks. This loss of granularity can impair the model's ability to detect subtle patterns or variations, which may be essential in high-precision tasks or domains where minimal differences are of great significance. Furthermore, determining appropriate criteria and thresholds for merging instances presents an additional challenge. Striking the right balance is crucial; a merging policy that is too liberal may lead to overgeneralization, while one that is too conservative may fail to effectively reduce redundancy.

Dataset distillation and condensation take a more synthetic approach, aiming to condense or summarize the original dataset into a smaller, yet information-dense, version that retains the essential characteristics of the original~\cite{liu2021dataset}. This approach not only reduces the dataset’s size but also provides a more efficient representation, allowing for faster computation without a substantial loss in predictive power.
In contrast, dataset pruning involves reducing the size of a dataset by removing less significant data points~\cite{wilson2000reduction}. This technique is employed to eliminate noise and redundant information that do not contribute meaningfully to the learning process, thus enhancing computational efficiency and, in some cases, improving the accuracy of data mining algorithms. In particular, data selection focuses on identifying and retaining the most relevant and informative data points within a dataset~\cite{liu2001instance}. Selection criteria may vary depending on the context, but the overarching goal is to preserve data that significantly contributes to model learning while discarding irrelevant or less informative instances. This method is particularly advantageous in contexts such as active or continual learning, where dynamic adaptation of the dataset is critical.
Among data selection methods, instance selection has emerged as a particularly promising technique, capable of achieving both greater efficiency and efficacy. Its primary strength lies in its ability to maintain, and in some cases, enhance the quality of the learning process by filtering out irrelevant or noisy data. In situations where datasets are cluttered with uninformative or redundant data, IS can effectively reduce this noise, leading to a streamlined dataset that improves the performance of learning models. For example, Liu and Motoda demonstrated how instance selection could enhance the performance of k-nearest neighbor classifiers by selectively pruning training data~\cite{liu2001instance}.
Additionally, instance selection methods are highly versatile and adaptable to a wide range of datasets and learning tasks. Whether applied to large-scale datasets or those with complex, imbalanced class distributions, IS techniques have proven effective in tailoring the dataset to the specific needs of the learning algorithm. This adaptability is particularly crucial in fields such as bioinformatics and medical imaging, where datasets tend to be both voluminous and intricate. However, despite its advantages, IS presents challenges in selecting the most appropriate criteria and algorithms for a given dataset and task. As outlined by the "no free lunch" theorem, no single IS method is universally optimal across all datasets and learning tasks~\cite{wolpert1997no}. Thus, selecting the most effective IS technique often requires careful consideration of the specific characteristics of the dataset and the learning problem at hand.

Elimination methods are fundamentally designed to enhance the quality of a dataset by removing noisy, irrelevant, or otherwise uninformative elements that do not contribute meaningfully to the task at hand~\cite{garcia2012survey}. The primary benefit of these methods lies in their capacity to cleanse the dataset, thereby enabling the development of more accurate and computationally efficient predictive models. By eliminating detrimental data points, elimination techniques can significantly improve the signal-to-noise ratio, providing greater clarity for complex analytical tasks. In scenarios where extraneous data is prevalent, this enhanced clarity can help mitigate the risks of misinterpretation or overfitting in machine learning models. However, the use of elimination methods comes with inherent risks. A major challenge is accurately distinguishing between data that is genuinely irrelevant or noisy and data that, while appearing superfluous, actually contains critical information for model training and prediction. In practice, elimination methods may inadvertently remove key instances, such as outliers or exceptions, that significantly contribute in identifying complex patterns or detecting anomalies. 

There are also techniques that eliminate samples based on affinity criteria. Among these, clustering algorithms, for example, are fundamental for revealing intrinsic structures within datasets by grouping similar instances~\cite{ester1996density}. These methods are particularly effective in identifying natural groupings, thereby aiding in data simplification while preserving essential characteristics. However, their effectiveness is highly contingent upon the specific clustering algorithm and parameters employed, as these can significantly influence the representation of data groupings and the identification of representative instances. For instance, methods such as affinity propagation and global density-based algorithms focus on the density of instances within feature space~\cite{frey2007clustering}. Affinity propagation, in particular, identifies exemplar data points to serve as cluster representatives, automatically determining the number of clusters without requiring predefined input. This feature enhances its ability to create high-quality exemplars and cluster structures. On the other hand, global density-based methods focus on the overall distribution of instances, retaining central points in dense regions. While these methods effectively balance data reduction with classification accuracy, they may struggle with datasets that have irregular density distributions, potentially overlooking important variations in the data. Typical or atypical instance selection methods aim to retain the most representative or unique instances within a dataset~\cite{liu2006outlier}. Typical instance selection captures the central tendency of the data, helping to understand predominant patterns. In contrast, atypical instance selection focuses on rare or unusual patterns, crucial for predictive tasks in domains where rare events are significant. However, this focus might introduce a bias towards outliers, potentially distorting the learning processes. Representativeness-based methods in instance selection focus on instances that best represent the entire dataset, creating a subset that reflects the complete diversity of the dataset~\cite{zhang2003selecting}. However, defining and quantifying "representativeness" can be challenging, often requiring complex calculations and potentially leading to biased subsets. Evidence-based methods use probabilistic measures for instance selection, providing a flexible and nuanced way to handle uncertainty in data~\cite{gong2021evidential}. These methods excel in scenarios with incomplete or ambiguous data but can be complex to implement and interpret, requiring significant computational resources. Locality-sensitive hashing (LSH) employs hash functions for efficient instance selection, particularly effective in high-dimensional data spaces~\cite{gionis1999similarity}. The scalability of LSH makes it suitable for big data applications, but its hashing process may miss subtle yet important differences between instances, crucial in scenarios where fine data variations significantly impact learning outcomes or model accuracy.

Cluster-oriented instance selection for classification problems represent advanced instance selection techniques, each with specific advantages and limitations. Cluster-oriented instance selection (CIS) seeks to enhance classification by identifying and selecting central instances within clusters, improving representativeness and reducing redundancy~\cite{saha2022cluster}. However, defining cluster parameters can significantly impact results and requires a deep understanding of data structure.
The Global Density-based Instance Selection (GDIS) algorithm introduces a relevance function alongside a global density function to determine the importance of instances within their k nearest neighbors, especially when class labels differ~\cite{malhat2020new}. While GDIS improves classification accuracy over traditional density-based approaches, it suffers from a reduced reduction rate. To address this, the Enhanced Global Density-based Instance Selection (EGDIS) algorithm incorporates an irrelevance function, which identifies instances in the k nearest neighbors that might lead to misclassification~\cite{malhat2020new}. By retaining only those instances likely to be misclassified and the densest instances, EGDIS enhances both the reduction rate and effectiveness compared to GDIS.

\section{RAZOR: the proposed approach}
With the rise of deep learning and the adoption of end-to-end learning protocols associated with increasingly complex deep architectures, the volume of data required for effective training—while avoiding overfitting—continues to grow, often at the expense of efficiency. The prevailing assumption is that more data leads to more information. However, this is not always the case. Shannon’s definition of entropy provides a framework for quantifying the amount of information contained in a dataset, from which the concept of data redundancy is derived. In essence, a dataset contains redundant data when certain information is repeated across multiple samples. Consequently, more data does not necessarily imply more information. In line with this understanding, a deep learning architecture will improve its performance with larger datasets only if the additional data is not redundant~\cite{sorscher2022beyond}.

To reduce redundancy and improve the training process's effectiveness and efficiency, we propose in this work an instance selection method that, like other known techniques, relies on clustering to group similar instances and select only a few representatives from each group. The role of the clustering algorithm is crucial, as the goal is to reduce the number of training instances by selecting the most representative ones. 

We propose a clustering algorithm based on the concept of representativeness of a sample with respect to the set of samples to which it belongs, where the representativeness is considered an estimate of the informational content that the sample contributes to its group. In this context, the concept of representativeness is closely linked to the notion of information and is thus defined in terms of the entropy concept introduced by Shannon. This approach allows for identifying and retaining only instances that significantly contribute to the variety and completeness of the information available in the dataset, thus supporting the effectiveness of the learning process.

The clustering algorithm must evaluate the representativeness of each sample within the set to effectively partition the entire set of instances into homogeneous clusters. This is an extremely onerous process whose cost increases proportionally with the size of the set of elements. This characteristic makes the algorithm's direct application to very large sets unfeasible.

Operating separately on smaller portions of the entire set is necessary to contain the clustering algorithm's complexity. However, this fragmentation can overlook the interactions between elements initially assigned to different subgroups, thus limiting the clustering process's effectiveness.

To overcome this limitation, an iterative clustering approach has been developed that alternates between two distinct phases: a cluster generation phase, which operates on reduced partitions of the set, and an aggregation phase, which combines the clusters formed in the previous phase to reconstruct the set of elements in a new configuration.

The alternation of these phases continues until cluster stability is reached. Only in this final phase an additional aggregation is performed, during which clusters deemed similar are merged according to a criterion specifically designed for this purpose. This strategy aims to maintain a balance between reducing computational complexity and preserving the integrity and representativeness of the information within the original set.

When the clustering is completed, the sample selection step is performed by analyzing the convex hull of each cluster to identify the most distinctive and representative points. More in detail, sample selection begins by calculating the Euclidean distance matrix between points within the cluster and identifying a central key point. Principal Component Analysis (PCA) is then applied to reduce the data dimensionality, facilitating the identification of the cluster's main features.
Subsequently, the convex hull is calculated, and the frequency with which each point appears as a vertex is determined, indicative of its geometric importance. Points with high frequencies are considered highly representative of the cluster. A predetermined fraction of these points is then selected for training. 
This approach ensures the diversity and statistical significance of the samples, thereby enhancing the robustness and effectiveness of the final model.

\begin{figure*}[ht]
\centering
\includegraphics[width=1.0\textwidth]{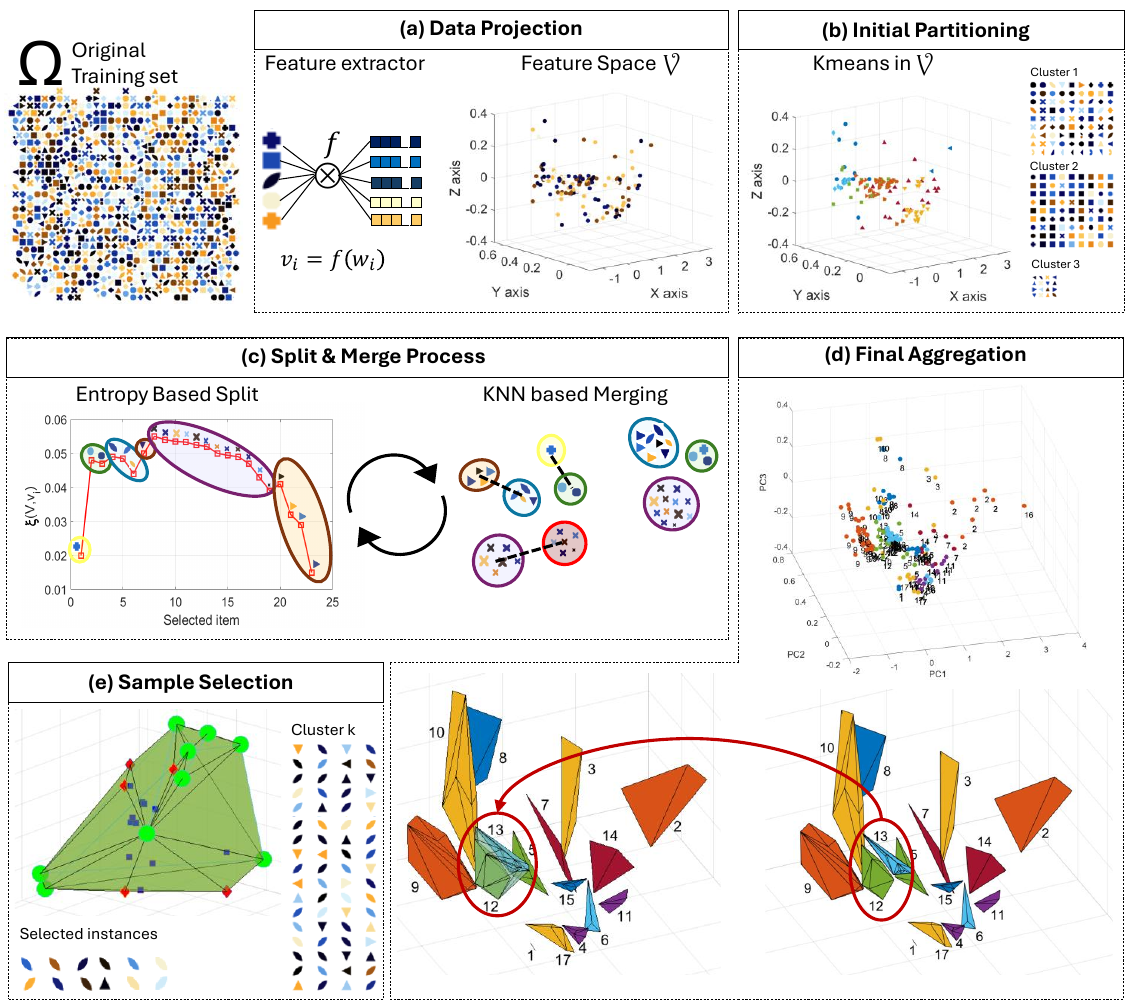}
\caption{Overview of RAZOR's instance selection algorithm based on entropy clustering. The process includes several key steps: (a) \textbf{Data Projection}: The original training set $\Omega$ is processed through a feature extractor to obtain feature vectors $v_i = f(\omega_i)$, which are then mapped into a feature space $\mathcal{V}$; (b) \textbf{Initial Partitioning}: The dataset $\mathcal{V}$ is partitioned using k-means clustering into initial clusters to ensure manageability; (c) \textbf{Split \& Merge Process}: An iterative process involving entropy-based splitting and KNN-based merging is employed to refine the clusters. The entropy-based split phase identifies representative samples while the KNN-based merge phase consolidates similar clusters; (d) \textbf{Final Aggregation}: The final set of clusters is obtained merging clusters according to the shape of the corresponding convex hulls; (e) \textbf{Sample Selection}: The convex hull of each final cluster is analyzed to identify the most distinctive and representative points, which are then selected for the final training set. Note that steps b-e do not involve the original objects in $\Omega$ but are carried out directly in $\mathcal{V}$. In (c) and (e), the parallel with the original objects is reported solely to clarify the processing procedure.}
\label{fig:architecture}
\end{figure*}

\subsection{Entropy-Based Clustering}
\label{subsec:EAC}
In the process of training a classifier, high-quality samples alone are insufficient for ensuring robustness against distortions and input variability. Therefore, maintaining a diverse set of training samples is essential to improve the classifier's ability to generalize effectively across various scenarios.
To address this issue, Shannon's entropy represents a powerful tool to evaluate the representativeness of a set of samples $\Omega$ for training a model. 
Indeed, by identifying groups of similar or redundant samples, entropy helps optimize the dataset by selecting diverse and representative samples.

To implement this approach, we adapt Shannon's definition of entropy to consider the relationships (similarities) among the elements in the original training set $\Omega$. 
This technique is completely generic and can be applied to training sets with data of any kind, provided that there exists a mapping function that associates to each sample a representation in a numeric feature space. Hence, we assume the existence of a feature extraction technique $f(\omega_i) = v_i$, which, given an element $\omega_i$ from the set $\Omega$, produces its embedding $v_i$ in an $m$-dimensional vector space. A similarity measure $d$ then assigns a scalar value $s_{i,j} = d(v_i, v_j)$ to each pair of elements $(v_i, v_j)$, where $s_{i,j}$ lies within the range $[0,1]$. If necessary, normalization ensures that the values of $d$ fall within this range. 

Entropy is used to evaluate the representativeness of the set $\mathcal{V}$, which consists of the embeddings derived from $\Omega$. Here, $s_{i,j}$ is interpreted as the probability that element $v_i$ conforms to $v_j$, with the sum of all $s_{i,j}$ values over $\mathcal{V}$ normalized to 1. The entropy of an element $v_j$ within the set $\mathcal{V}$ is defined as:

\begin{equation}
H(\mathcal{V}, v_j) = - \frac{1}{\log_2(|\mathcal{V}|)} \sum_{i=1}^{|\mathcal{V}|} s_{i,j} \log_2(s_{i,j}), 
\end{equation}

where the normalization factor $\frac{1}{\log_2(|\mathcal{V}|)}$ ensures that the values range between 0 and 1, regardless of the size of $\mathcal{I}$.

To measure the diversity of the entire set $\mathcal{V}$, we calculate:

\begin{equation}
H(\mathcal{V}) = - \frac{1}{\log_2(|S^+|)} \sum_{s_{i,j} \in Q} s_{i,j} \log_2(s_{i,j}), 
\end{equation}

where $S^+$ is the set of all scores $s_{i,j} > 0$ computed between pairs $(v_i, v_j)$ in $\mathcal{V}$. 

Entropy helps select a subset of representative samples by evaluating the effect of removing each element on the overall entropy of the set. 
For each element $v_i \in \mathcal{V}$, we calculate the entropy difference $\xi(\mathcal{V}, v_i) = H(\mathcal{V}) - H(\mathcal{V} \setminus \{v_i\})$. 

This selection process tends to show a stepwise progression in the function $\xi$, where small variations are followed by significant spikes when a distinctly different sample is chosen. This behavior is modeled by a parabolic trend due to larger initial variations in the differential, stabilizing and reversing as the number of remaining samples decreases. This characteristic behavior is clearly illustrated in Figure~\ref{fig:architecture} (c).

Using the function $\xi$, we stratify the set $\mathcal{V}$ into smaller, homogeneous subsets or clusters. The procedure begins with a single cluster $S_1$, incorporating the most representative element. As the algorithm progresses, elements are added to existing clusters based on their similarity (decremental model in $\xi$) or start new clusters if distinctly different (incremental model in $\xi$).
Each time $\xi$ increases, a new cluster is generated, whereas points for which the function decreases are added to the last generated cluster. At the end, a complete partition of the data is obtained, represented by the colored ovals in~\ref{fig:architecture} (c).

This iterative process continues until all elements in $\mathcal{V}$ are evaluated and categorized into clusters. This approach ensures that the resulting clusters are both representative and diverse, enhancing the robustness of the training dataset for the classifier.

Due to its complexity, the entropy-based clustering algorithm cannot be directly applied to very large sets of samples. To address this limitation, we designed an iterative split and merge process that involves a partitioning phase, a clustering phase on individual partitions, and an aggregation phase of the produced clusters. 

\subsection{Data partitioning}
Since greater consistency of samples in the initial partitions speeds up the convergence of the entire clustering process, we decided to use k-means to generate the initial set of partitions. 
However, when the initial number of samples is in the hundreds of thousands and the dimensionality of the points is also in the thousands, even the k-means algorithm can be inefficient if applied to the entire set.
To ensure high scalability, we set an upper limit $\text{N}_{kmeans}$ on the size of the initial partitions to which the k-means algorithm is applied. 
 If $|\mathcal{V}| > \text{N}_{kmeans}$, the set $\mathcal{V}$ undergoes a simple partitioning step splitting $\mathcal{V}$ into $|\mathcal{V}|/\text{N}_{kmeans}$ subgroups $Q_i$, such that $Q_i \bigcap Q_j=\emptyset \ \forall i\neq j$ and $\bigcup_i Q_i = \mathcal{V}$. 

On each partition $Q_i$, a k-means clustering step is performed to produce $|Q_i|/\text{N}_{entcls}$ partitions $S_j$, where $\text{N}_{entcls}$ is the maximum number of samples for which the entropy-based clustering algorithm can efficiently be applied. 
If after the k-means step, there are still some partitions with sizes greater than $\text{N}_{entcls}$, then these are further subdivided using a simple split process.

The set $\mathcal{S}$ of all partitions $S_i$ with $i=1,...,|\mathcal{S}|$ such that $\mathcal{V}=\bigcup_{i}S_i$ and $S_i \bigcap S_j = \emptyset$ $\forall i \neq j$, constitutes an initial subdivision of the dataset $\mathcal{V}$, ensuring that no partition has more elements than $\text{N}_{entcls}$ and forms the initial set of clusters for the iterative split and merge process cycle.

Then, for each $S_i$ in $\mathcal{S}$ the centroid $c_i$ is calculated with a standard Euclidean distance.  
A centroid is the midpoint of all data points within a cluster, serving as a representative marker that typifies the average position of points in the cluster. These centroids form a new set $C$, capturing the central points of all clusters in $\mathcal{S}$. The sets $S$ and $C$ are the initial input to the following entropy-based clustering steps.

\subsection{Iterative split and merge process}
The algorithm proceeds iteratively, alternating clustering phases and aggregating phases, with its convergence automatically evaluated, up to a predefined maximum of $Max_{iter}$ iterations. 
At the start of each iteration, a new set of clusters and centroids, denoted as $\mathcal{S}'$ and $C'$, respectively, are initialized as empty. Each $S_i$ in $\mathcal{S}$ is given as input to the entropy clustering algorithm, which further partitions $S_i$ into a resulting set of new clusters that are inserted into $\mathcal{S}'$, while the corresponding centroids are inserted into $C'$.

Since the subgroups $S_i$ are processed independently, but the goal is to properly cluster the whole set $\mathcal{V}$, an aggregation phase is essential. 
In this step, the clusters produced by the independent applications of the entropy-based clustering algorithm collected in $\mathcal{S}'$ are reassembled into a single set of clusters comprising all elements of $\mathcal{V}$. This aggregation ensures that, despite the separate processing of the subgroups, the final clustering result is representative of the overall structure and distribution within the entire dataset $\mathcal{V}$. It bridges the gap between localized clustering in the subgroups and a globally coherent clustering solution.

If the current iteration is not the last one, i.e., $Max_{iter}$, the aggregation step is performed as follows. A \textit{KNN} search is applied to $C'$, seeking the nearest centroid $c_j$ for each centroid $c_i$ in $C'$, excluding itself. The corresponding clusters $S_i$ and $S_j$ are merged if neither has been previously merged with others. After the merging process, a check is made to control whether the union of two clusters has produced a cluster whose sizes exceed the $\text{N}_{entcls}$ limit.
If this is the case, the cluster is subdivided into the minimum number of new clusters necessary to ensure that none exceeds the $\text{N}_{entcls}$ limit. After determining the new set of clusters $\mathcal{S}'$, the corresponding centroids $C'$ are recalculated. 

After the aggregation process, the algorithm evaluates convergence by comparing the quality of the clusters $C$ from the previous iteration with those of the clusters $C'$ obtained from the current iteration. 

Given the high number of clusters and the elements within them that need to be managed, and the requirement to evaluate convergence at each iteration, the chosen cluster similarity measure must be both robust and efficient. To this end, the nearest Intersection over Union (nIOU) has been defined to compare the similarity between two sets of clusterings \(C_1\) and \(C_2\).

First, for each cluster \(c_i\) in \(C_2\), the function identifies the nearest cluster \(c_h\) in \(C_1\) using the k-nearest neighbors (kNN) search. The Intersection over Union (IoU) between each cluster in \(C_2\) and its nearest cluster in \(C_1\) is then computed. The IoU is calculated as the ratio of the intersection of the two clusters' elements to their union. This process is then repeated in the opposite direction: for each cluster \(c_j\) in \(C_1\), the nearest cluster \(c_k\) in \(C_2\) is identified, and the IoU is computed similarly. The IoU scores are averaged over all clusters:

\begin{equation}
\text{AVG}_{\text{overlap}} = \frac{1}{2} \left( \frac{\sum_{i=1}^{|C_2|} \text{IoU}(c_i, c_k)}{n_2} + \frac{\sum_{j=1}^{|C_1|} \text{IoU}(c_j, c_h)}{n_1} \right).
\end{equation}

To address the imbalance in the number of clusters between \(C_1\) and \(C_2\), a penalty term is introduced. This term is based on the ratio of the total number of clusters, aiming to normalize the measure to ensure it reflects both the overlap between clusters and the disparity in cluster counts:

\begin{equation}
\text{penalty} = \frac{n_1 + n_2}{2 \cdot \max(n_1, n_2)}. 
\end{equation}

The final nIOU score is the average of \(\text{AVG}_{\text{overlap}}\) and \(\text{penalty}\). The value of this measure varies in the range \([0,1]\), where 0 represents maximum diversity and 1 represents a perfect match between the two clusterings.

The convergence measure $d_{conv}(C, C') = 1 - \text{nIOU}(C, C')$ is then computed. This convergence measure captures the dissimilarity between the clusterings of two successive iterations, with a lower value indicating higher similarity and convergence.

If the convergence measure is below a threshold $\epsilon$, the algorithm stops iterating. Otherwise, it proceeds to the next iteration, updating $C$ with $C'$ and $\mathcal{S}$ with $\mathcal{S}'$. The process is repeated until convergence or reaching the maximum number of iterations $Max_{iter}$.

\subsection{Final Aggregation}
A final cluster aggregation step aims to further refine the set of clusters obtained in the previous clustering phase. This step reduces the total number of clusters, particularly merging those previously separated due to their size exceeding the threshold $\text{N}_{entcls}$.

For clusters to be good candidates for merging, they need to be close, a condition verified by considering the distance between their centroids. Clusters with distant centroids can be discarded early in the merging process. This consideration reduces the computational cost, as a KNN search can retrieve a set $K$ of candidate neighbors for merging each cluster.

It can happen that multiple clusters $S_i$ have the same closest cluster $S_j$, although with different distances. To choose the best pairing for merging, a compatibility score is defined. The aggregation process is organized in two steps: first, identify all potential pairs of clusters for merging and calculate their compatibility score; second, sort the pairs based on the score and proceed with the actual merging, ensuring each cluster is merged only once and with the best remaining candidate.

The compatibility criterion is defined to decide whether a pair of clusters can be merged. Given clusters $S_i$ and $S_j$, the set $S_{i,j}=S_i \cup S_j$ is constructed, and Principal Component Analysis (PCA) is applied to project the elements into a 3D space. The score is calculated using the formula:

\begin{equation}
\phi_{i,j} =
\begin{cases}
1 & \text{if } |S_i| < 3 \text{ or } |S_j| < 3 \\
\frac{vch_i + vch_j}{vch_{i,j}} & \text{otherwise}
\end{cases},
\end{equation}

where $vch_i$, $vch_j$, and $vch_{i,j}$ are the volumes of the convex hulls of the projected points in 3D for $S_i$, $S_j$, and $S_{i,j}$, respectively.

This aggregation process is iterative and continues until no further merges are possible. For each cluster $S_i$, the nearest neighbor $S_j$ is retrieved, and the fusion score is calculated, constructing a list $L=\{(i,j,\phi_{i,j})\}$. The list $L$ is sorted in descending order of $\phi_{i,j}$. Starting from the top of the list $L$, and given a threshold $th_{\phi}$, if the pair of clusters $(S_i, S_j) \in L$ has a value of $\phi_{i,j} \geq th_{\phi}$, the two clusters are merged, and any other pairs in $L$ involving $S_i$ or $S_j$ are ignored. At the end of the process, the centroids of the newly obtained clusters are recalculated, and the procedure repeats until no further merges are made during the iteration. The final aggregation procedure returns the new set of clusters $S$ and their respective centroids $C$.

\subsection{Sample Selection}

Once the clustering procedure is concluded, the next important step is to choose the most representative samples. This selection procedure is designed to ensure that the samples chosen for classifier training are both representative of the various characteristics present in each cluster and sufficiently diverse to provide effective and balanced training.

We propose a selection step based on convex hull analysis because it allows for the capture of the variety and diversity of data within each cluster.
The convex hull, defined as the smallest convex polyhedron containing all the points in a set, is an effective method for identifying those points representing the cluster's extreme spatial characteristics.
The idea behind this technique is that the points forming the convex hull are the most distinctive, being at the "frontiers" of the cluster. These points are typically more informative as they represent the boundary cases of the cluster's characteristics, ensuring that data diversity is preserved.

Sorting the vertices of the convex hull based on the frequency of their occurrences provides an additional criterion for selecting representative samples. The frequency with which a point appears as a vertex in a convex hull of reduced-dimensional subspaces indicates its geometric importance in the cluster. Points with high frequencies are more likely to represent distinctive characteristics of the cluster.

This approach allows for the selection of samples that not only cover a wide range of characteristics but are also statistically significant, thus enhancing the robustness and effectiveness of the learning model. Using this methodology, the algorithm forms a training set that is representative of the entire data space, minimizing the risk of overfitting and ensuring good generalization of the classifier on unseen data.

The sample selection elaborates the set of clusters $\mathcal{S} = \{S_1, S_2, \ldots, S_n\}$ resulting from the final aggregation step aiming to select a predetermined percentage $\tau$ of training instances from each cluster to form a balanced and representative training set.

For each cluster $S_i \in \mathcal{S}$, the following steps are executed: the data points belonging to the cluster, denoted as $\mathcal{V}_{S_i}$, are considered. The distance matrix $\mathbf{D}_i$ is calculated, where each element $D_{i, j}$ represents the Euclidean distance between points $v_i$ and $v_j$ within the cluster $S_i$:

\begin{equation}
    D_{i,j} = \| v_i - v_j \|,
\end{equation}

with $v_i, v_j \in \mathcal{V}_{S_i}$.
    
The first key element $s_k$ in the cluster $S_i$ is identified by minimizing the sum of the mean and standard deviation of the distances of each point relative to the other points in the cluster:

\begin{equation}
    k = \underset{j}{\mathrm{argmin}} \left( \mathrm{mean}(D_{i, j}) + \mathrm{std}(D_{i, j}) \right).
\end{equation}
    
Principal Component Analysis (PCA) is applied to the points in the cluster $\mathcal{V}_{S_i}$, obtaining the component scores $Z_{S_i}$:

\begin{equation}
    Z_{S_i} = V_{S_i} W,
\end{equation}
where $W$ is the matrix of principal components. The top $m$ principal components are selected to reduce the dimensionality of the data. The convex hull $H$ of the reduced points is calculated, and the histogram of occurrences $h$ of the points in the convex hull is determined.

The points are sorted in descending order of occurrences $h$, and the first $N_{samp}$ instances are selected, where $N_{samp}$ is calculated as a fraction $\tau$ of the total number of instances in $S_i$, that is:

\begin{equation}
    N_{samp} = \lceil \tau \times |v_{s_i}| \rceil.
\end{equation}

At the end of the process for each cluster, the selected samples are collected into an output set denoted as $\Omega_{\mathcal{S}} \subset \Omega$. This set represents the final subset of instances chosen for classifier training, ensuring that the samples are representative and diverse, enhancing the effectiveness of the training process.

\subsection{Computational Complexity Analysis}
\label{subsec:ACC}
Given that RAZOR is compared to several state-of-the-art techniques, namely CIS~\cite{saha2022cluster}, GDIS~\cite{malhat2020new}, EGDIS~\cite{malhat2020new}, and ENRICH~\cite{chinn2021enrich}, it is essential to consider the computational complexity of each method to ensure a comprehensive evaluation of their performance when applied to a dataset of $n$ input samples that have been projected into an $m$-dimensional feature space.
This analysis provides a theoretical foundation to support the empirical findings, offering valuable insights into the efficiency and scalability of the algorithms. Understanding the computational demands of each approach helps clarify their performance in terms of time and resource consumption. In particular, RAZOR's complexity arises from multiple phases—partitioning, clustering, aggregation, and sample selection—each of which contributes to the overall computational cost.
The computational complexity of the RAZOR algorithm is influenced by the clustering, aggregation, and sample selection phases. Initially, the dataset is partitioned into subgroups, and each subgroup undergoes k-means clustering, resulting in a complexity of \(O(n \cdot k \cdot m)\), where \(n\) is the number of samples, \(k\) is the number of clusters, and \(m\) is the number of features. RAZOR is parallelizable, allowing the k-means clustering and entropy-based clustering steps to be distributed across \(N_{\text{cores}}\) processors. Thus, the iterative clustering phase, which relies on entropy-based clustering, has a complexity of
\begin{equation}
O\left(\textit{maxiter} \cdot\frac{ n \cdot ( N^3_{\text{entcls}} +  \log(n))}{N_{\text{cores}}}  \right),
\end{equation}
with \(N_{\text{entcls}}\) representing the maximum number of samples allowed for the entropy-based clustering step.

The CIS method employs K-means clustering to select representative instances from each cluster. Its computational complexity arises from performing the clustering operation, which is $O(n \cdot m)$, followed by calculating the Euclidean distances for each instance within a cluster, with a complexity of $O(n_i \cdot m)$ for each cluster $i$. Sorting these distances adds an additional complexity of $O(n_i \log n_i)$. Summing over all clusters, the total complexity for CIS becomes $O(n \cdot m + n \log n)$. The GDIS algorithm selects instances based on their density and relevance to their $k$-nearest neighbors. The dominant complexity in this case stems from computing the $k$-nearest neighbors for each instance, which leads to a complexity of $O(n^2 \cdot m)$. Additional steps, such as relevance and density calculations, maintain the overall complexity of $O(n^2 \cdot m)$. EGDIS, which operates similarly to GDIS, adds irrelevance calculations to the selection process. Like GDIS, the complexity is driven by the computation of $k$-nearest neighbors, as well as the density and irrelevance calculations. This yields the same overall complexity as GDIS, i.e., $O(n^2 \cdot m)$. Finally, the ENRICH method works by iteratively selecting diverse instances based on cosine distance calculations between selected and unselected instances. The main computational burden in this method comes from these distance calculations, leading to a total complexity of $O(\tau \cdot n^3 \cdot m)$, where $\tau$ is the selection rate that determines the portion of instances to be selected.
RAZOR stands out from other methods due to its lower computational complexity and superior scalability. Unlike the other methods, which often involve more computationally intensive operations, RAZOR maintains a simpler complexity structure. Furthermore, its ability to parallelize clustering and selection processes across multiple cores makes it highly scalable, efficiently managing large datasets while reducing execution time, which is a distinct advantage in big data environments.

\section{Experiments and Results}
This section describes the experimental aspects of the RAZOR system on both synthetic and real datasets, as well as a comparison of its performance with recent techniques from the literature. Specifically, it illustrates the synthetic data generation process and the characteristics of the real datasets. It then describes the process of projecting the data into feature space, the parameter configuration values adopted, and the individual experiments conducted. The effectiveness of the different methods is assessed using appropriate measures chosen based on the type of data or task involved and the statistical validity of the results is also verified with the Wilcoxon signed-rank test. Efficiency is evaluated by measuring execution times on a machine with the following configuration, which was used for all tests: 13th Gen Intel(R) Core(TM) i7-13700H 2.40 GHz, 16.0 GB of RAM, Windows 11 Home 64-bit, and an NVIDIA GeForce RTX 4060 Laptop GPU with 16 Gb of RAM.

\subsection{Pre-set Parameters}
RAZOR is designed to be a robust, versatile, and scalable algorithm with a minimal number of parameters to set. The main parameters include the number of clusters for the initial k-means clustering step, the maximum number of elements for the input of the entropy-based clustering, the maximum number of iterations for the iterative process, and the convergence threshold.

The number of clusters for the initial k-means clustering (\(k\)) is a key factor for generating the initial set of partitions. In our experiments, this parameter is set to $1000$ to ensure a comprehensive and manageable initial partitioning of the dataset. Another key parameter, \(N_{entcls}\), defines the number of elements for the entropy-based clustering. We fixed \(N_{entcls}\) at $100$, balancing the need to assess the homogeneity of the set with the computational efficiency of the algorithm.

The iterative process is controlled by the maximum number of iterations (\(Max_{iter}\)), which we experimentally set to $10$. This value was sufficient to achieve convergence in all test cases, making it a practical upper limit. Finally, the convergence threshold (\(\epsilon\)) determines when the iterative process should stop based on the convergence measure. We set this threshold to $0.05$, corresponding to an nIOU value of $0.95$. This value ensures a stable clustering result, aligning with the goals of the iterative entropy-based clustering process.

Additionally, the PCA dimension for the final aggregation step is set to \(3\), which facilitates the merging process by projecting data into a more manageable space. The threshold for merging clusters, denoted as \(th_{\phi}\), is \(0.9\), ensuring that only highly similar clusters are combined. For the sample selection step, the PCA dimension is set to \(8\), allowing for a detailed capture of the cluster's characteristics while maintaining computational efficiency.

\subsection{Data Projection and the similarity measure}

As explained in Subsection~\ref{subsec:EAC}, the elements in the original training set need to be projected into a feature space $\mathcal{V}$ using a feature extractor $f$. For the images used in the experiments within this paper, the ResNet50 network architecture was chosen. The output from the last average pooling layer of the ResNet50 network serves as the feature vector associated with the images. This feature vector consists of 1024 real values in the range $(-\infty, \infty)$, thereby generating a 1024-dimensional feature space $\mathcal{V}$, denoted by the parameter $m$ in the method description.
Although it is not a parameter in the strict sense, the similarity measure $d$ described in Section~\ref{subsec:EAC} can vary, provided that its values are kept within the range \([0,1]\). In the specific case, \( d(v_i, v_j) \) is a normalized version of the correlation similarity and is defined as:

\begin{equation}
s_{i,j} = \frac{1}{2} \cdot \left(\frac{(v_i - \bar{v_i}) \cdot (v_j - \bar{v_j})}{\| v_i - \bar{v_i} \|_2 \| v_j - \bar{v_j} \|_2}+1\right)
\end{equation}

where \(\bar{v_i}\) and \(\bar{v_j}\) are the mean values of the feature vectors \(v_i \in \mathcal{V}\) and \(v_j \in \mathcal{V}\), respectively, \(\cdot\) represents the dot product, and \(\|\cdot\|_2\) represents the Euclidean norm.

\subsection{Evaluation Metrics}
The experiments on synthetic data aim to estimate the ability of the clustering algorithm to correctly identify the number of classes and their respective elements in an unsupervised manner. Since the data are synthetically generated, a ground truth of clusters and their associated elements is available. The measure is calculated between the ground truth and the clustering generated by the iterative entropy-based clustering algorithm. To evaluate the performance, the Adjusted Mutual Information (AMI) was adopted~\cite{vinh2010information}. This metric is chosen as it measures the agreement of the two clusterings by considering the mutual information and correcting it for chance, thus providing a normalized measure that accounts for the expected similarity between clusterings. 

As regards the experiments on real data, the evaluation metrics were selected to match the different types of image analysis tasks, including classification and segmentation (both binary and multi-class). For classification tasks, Accuracy and F1-score were chosen. 

For binary and multi-class semantic segmentation tasks, the metrics used were F1-score, Intersection over Union (IoU), Dice coefficient, and Overlap Rate. These metrics are computed for each class individually and then averaged over all classes to obtain the final performance measures.

The IoU for a specific class \(c\) is defined as:

\begin{equation}
\text{IoU}_c = \frac{TP_c}{TP_c + FP_c + FN_c}
\end{equation}

The Dice coefficient for a specific class \(c\) is calculated as:

\begin{equation}
\text{Dice}_c = \frac{2 \times TP_c}{2 \times TP_c + FP_c + FN_c}
\end{equation}

The Overlap Rate for a specific class \(c\) is defined as:

\begin{equation}
\text{OverlapRate}_c = \frac{\sum_{i,j} (I_{i,j} == c \land GT_{i,j} == c)}{\sum_{i,j} (GT_{i,j} == c)}
\end{equation}

where \(I_{i,j}\) and \(GT_{i,j}\) represent the predicted and ground truth labels for the pixel at position \((i,j)\), respectively.

The performance metrics for the multi-class problem are averaged over all classes.

The Wilcoxon signed-rank test was employed to validate the statistical significance of the results. This non-parametric statistical hypothesis test compares two related samples and evaluates whether their population mean ranks differ, making it appropriate for assessing paired differences.

\subsection{Synthetic Data}
In the proposed experimental framework, the objective is to create a dataset with a specific number of clusters, $N_c$, each containing a number of samples $n_s$, represented by vectors of dimension $m$. The variability within each cluster is controlled by $\mu$, which determines the dispersion of data points around their respective cluster centers. The first step involves the critical task of defining the cluster centers, $c_i$, within an $N_c \times m$ dimensional space. 

In our experiments, a data matrix $X$ with dimensions $N_c \times n_s$ for $m$ is constructed to store the generated samples. Initially, this matrix is filled with zeros, establishing a structured base for the dataset. Data points for each cluster are generated by sampling from a multivariate normal distribution centered on each cluster center, $c_i$, with a covariance of $\mu$ times the identity matrix of dimension $m$. This method ensures that the distribution of points reflects the desired variability and isotropy for each cluster.

After generating the data points for all clusters, they are methodically inserted into the data matrix $X$, preserving the sequence of clusters. To infuse randomness and prevent any potential bias in the generation process, the rows of $X$ are randomized. This shuffling ensures that the dataset is free from sequential patterns that could unintentionally influence the results of clustering analyses.

\begin{figure}[ht]
\centering
\includegraphics[width=0.48\textwidth]{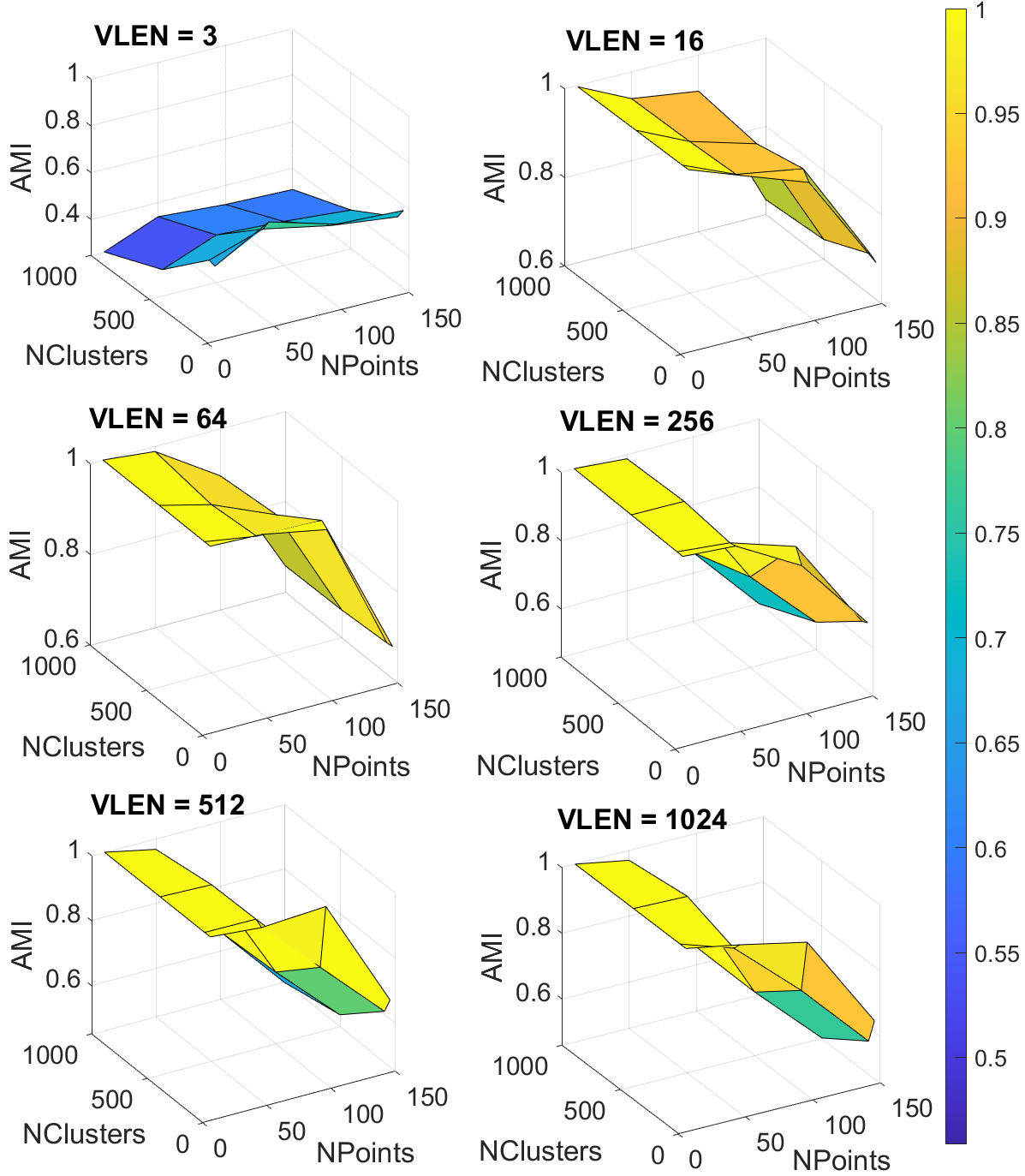}
\caption{The performance of the RAZOR algorithm, as measured by Adjusted Mutual Information (AMI). Six subplots depict different VLEN (vector length) values. The AMI values are plotted against the number of clusters (NClusters) and the number of points per cluster (NPoints). Each subplot provides a color-coded AMI gradient, with yellow representing higher AMI values and blue representing lower values.}
\label{fig:ami}
\end{figure}

Figure~\ref{fig:ami} presents the Adjusted Mutual Information (AMI) values achieved by the RAZOR algorithm under different synthetic data configurations. The results underline the capability of RAZOR to adapt to variations in vector length (VLEN), number of clusters (NClusters), and number of points per cluster (NPoints). As the vector length increases from 3 to 1024, AMI values noticeably improve across all conditions of NClusters and NPoints, indicating that longer vector lengths enhance the algorithm's ability to capture detailed data features, leading to more accurate clustering. This aligns with the understanding that higher-dimensional feature spaces improve the separation of complex data structures, albeit with potential challenges like the curse of dimensionality.

RAZOR shows consistent performance with larger numbers of clusters (500 to 1000), suggesting it effectively maintains clustering quality as the number of clusters increases. However, a slight drop in AMI at lower cluster counts (fewer than 100) for certain VLEN values indicates that the algorithm might struggle to distinguish clusters adequately when there are fewer distinct groupings. Moreover, an increase in NPoints generally correlates with higher AMI scores, especially at higher VLEN values, implying that RAZOR benefits from larger datasets, which refine cluster boundaries and enhance clustering stability.

The algorithm's ability to maintain high AMI values across various settings highlights its scalability and robustness, effectively handling variations in data density and complexity without significant degradation in clustering quality. 

\begin{figure*}[ht]
\centering
\includegraphics[width=0.48\textwidth]{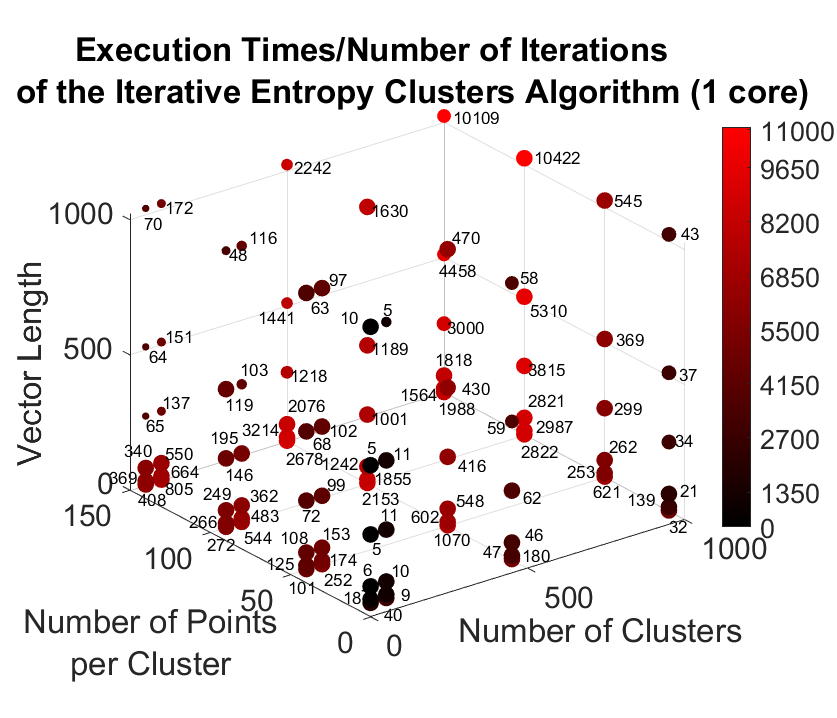}
\includegraphics[width=0.48\textwidth]{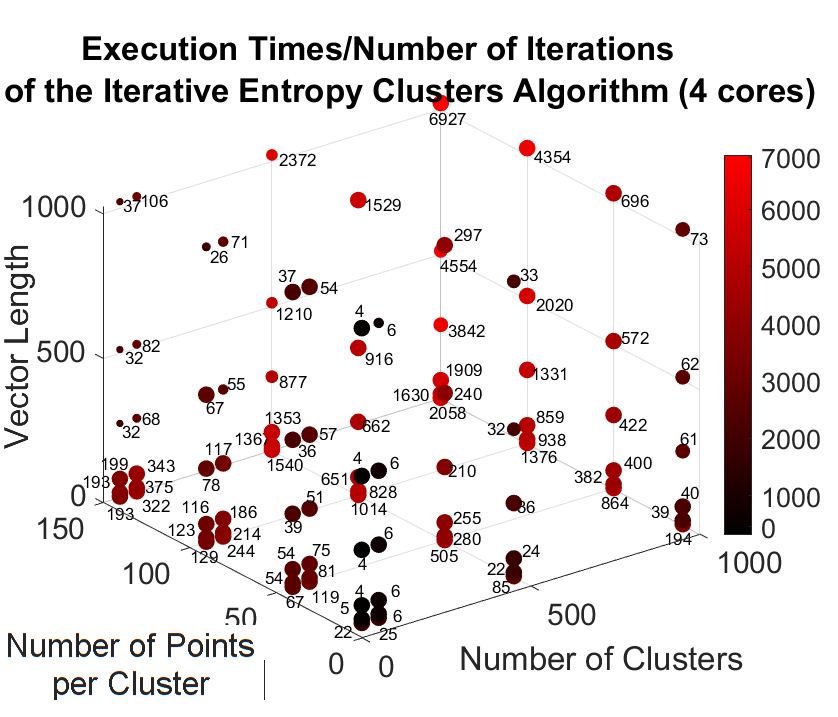}
\caption{The execution times and number of iterations required by the Iterative Entropy Clusters Algorithm under varying conditions, with results shown for single-core (left) and four-core (right) processing. The three axes represent the number of points per cluster, the number of clusters, and the vector length, respectively. The size of each point represents the number of iterations required to converge (from 1 to $maxiter$), while the brightness of the points corresponds to the execution time, which is also displayed numerically next to the points in seconds. }
\label{fig:time_iter}
\end{figure*}

Figure~\ref{fig:time_iter} illustrates the execution times and the number of iterations required for the Iterative Entropy Clusters Algorithm under varying conditions, with the number of points per cluster, number of clusters, and vector length represented on the three axes. The size of the dots reflects the number of iterations needed to reach convergence (from 1 to $maxiter$), where larger dots indicate a higher number of iterations. Concurrently, the brightness of the dots corresponds to the execution time, with brighter dots representing longer times, which are numerically annotated next to each point in seconds. The left side of the figure displays the results when a single core is used, while the right side shows the results when four cores are utilized.

The plots in Figure~\ref{fig:time_iter} reveal several key insights. As the number of points per cluster increases, there is a significant rise in execution time, whereas the number of clusters has a less pronounced effect. Furthermore, as the vector length increases, the algorithm tends to benefit from the higher dimensionality of the feature space, frequently achieving convergence with fewer iterations than the maximum allowed ($maxiter$), as indicated by the smaller dot sizes. When comparing the two graphs, it becomes evident that parallelization across multiple cores has a more substantial impact when the number of clusters is relatively low. This observation can be attributed to the overhead associated with managing parallel tasks across different cores, which becomes more significant as the number of clusters increases.

\subsection{Real Case Studies}
These experiments are mainly aimed at assessing the effectiveness of the sample selector that operates on the final clusters produced by entropy-based clustering. 
The goal is to validate the method's ability to extract a representative subset of training instances that ensures the training process's efficacy while simultaneously improving its efficiency.
The chosen application context for the experiments is the training of CNN networks applied to images
and, more specifically, to medical images. The latter were selected for the validation of RAZOR due to several compelling reasons. 
Since most imaging modalities have become digital, with ever-increasing resolutions, medical image processing must tackle the challenges associated with big data. However, many medical images are highly similar, introducing significant redundancy when used as training samples for tasks such as preprocessing, segmentation, and classification. Additionally, the suitable dataset size, the selection of appropriate training samples, and the data imbalance problem are critical factors that significantly impact the learning process in image analysis tasks. The main challenges in automated medical image analysis include the limited availability of annotated data and the requirement for large training datasets. RAZOR is designed to be a scalable algorithm that works efficiently with large volumes of data. It selects a representative subset of instances using entropy to minimize redundancy and operates without any prior knowledge of instance labels, choosing samples based on automatically identified groups, thereby not being influenced by potential label imbalances in the original dataset. For these reasons, the field of medical images represents the most suitable benchmark for evaluating RAZOR's capabilities.

In this work, we compare the performance of the RAZOR system with several state-of-the-art instance selection techniques, including CIS, GDIS, and EGDIS. These methods are general and, like RAZOR, can be applied to any type of data. Given that the domain of medical imaging was chosen for the experiments, an additional method, ENRICH, is included in the comparisons. Unlike the previous methods, ENRICH is specifically designed for instance selection applied to medical images.

In order to provide a comprehensive and diversified evaluation of RAZOR's effects on CNN training, three different types of applications have been considered: image classification, binary segmentation, and multiclass pixel classification (regression). For image classification, histopathological images were selected as the benchmark. For binary and multi-class semantic segmentation tasks, MRI image volumes were chosen as the testbed. 

In order to guarantee a comprehensive benchmark for evaluating the effectiveness of instance selection methods across different network complexities, three distinct architectures were selected as benchmarks that are AlexNet~\cite{krizhevsky2012imagenet}, VGG16~\cite{simonyan2014very}, and ResNet50~\cite{he2016deep}. These architectures provide a diverse range of structural characteristics that make them suitable for assessing how instance selection affects neural networks. Indeed, AlexNet, known for its relatively simple architecture with fewer layers and parameters, provides insight into how instance selection can impact networks with limited capacity for feature representation. VGGNet, with its deeper architecture and use of small convolutional filters, serves as a more complex network, allowing us to observe the effect of instance selection on models with greater depth and increased parameterization. ResNet50, characterized by its innovative residual connections, adds another layer of complexity, illustrating the potential benefits of instance selection in networks that mitigate the vanishing gradient problem and enable the training of very deep networks. 

\subsubsection{Histopathological Images}
Whole Slide Images (WSIs) are high-resolution images that are often divided into smaller regions for processing, resulting in a vast number of sub-images. This partitioning is necessary to manage the high resolution and data volume inherent in WSIs but also contributes to a large volume of data that must be managed effectively. The nature of histopathological images presents significant challenges for multi-class classification tasks, as the visual differences between classes can be subtle and complex. 
\begin{table}[h!]
\centering
\caption{Time (in hours) required by different instance selection methods to select a portion $k$ of samples from the entire training set. The number of cores used by RAZOR is indicated in parentheses.}
\begin{tabular}{l|*{6}{c}}
\hline
       & \multicolumn{6}{c}{K} \\
\cline{2-7}
       & 0.01 & 0.05 & 0.10 & 0.15 & 0.20 & 0.25 \cr 
\hline \hline
RAZOR(1) & 3.76 & 3.76 & 3.76 & 3.76 & 3.76 & 3.76 \cr 
\hline
RAZOR(4) & 1.09 & 1.09 & 1.09 & 1.09 & 1.09 & 1.09 \cr 
\hline
ENRICH         & 2.80 & 14.42 & 28.85 & 43.21 & 57.63 & 72.35 \cr 
\hline
CIS            & 0.0043 & 0.0099 & 0.0046 & 0.0036 & 0.0066 & 0.0101 \cr 
\hline
GDIS           & - & - & - & 18.20 & - & - \cr 
\hline
EGDIS          & - & - & - & 12.09 & - & - \cr 
\hline
\end{tabular}
\label{tab:results-histo-seltimes}
\end{table}
In this study, two distinct datasets, "NCT-CRC-HE-100K" and "CRC-VAL-HE-7K", were used. The "NCT-CRC-HE-100K" dataset comprises 100,000 non-overlapping image patches obtained from hematoxylin and eosin (H\&E) stained histological sections of normal and cancerous colorectal tissues. The images, sized 224x224 pixels (px) with a resolution of 0.5 microns per pixel (MPP), were chromatically normalized using the Macenko method. The tissue samples belong to different classes: adipose tissue (ADI), background (BACK), debris (DEB), lymphocytes (LYM), mucus (MUC), smooth muscle (MUS), normal colon mucosa (NORM), cancer-associated stroma (STR), colorectal adenocarcinoma epithelium (TUM). These images were manually extracted from 86 human cancer tissue slides stained with H\&E, derived from formalin-fixed paraffin-embedded (FFPE) samples from the NCT biobank (National Center for Tumor Diseases, Heidelberg, Germany) and the UMM pathological archive (University Medical Center Mannheim, Mannheim, Germany). The samples included slides of primary CRC tumors and CRC liver metastasis tissues, while normal tissue classes were supplemented with non-tumor regions from gastrectomy samples to increase variability.
\begin{table*}[ht]
\centering
\caption{Balance in class selection (mean and variance of the percentage of extracted subsamples from each class).}
\begin{tabular}{l|cc|cc|cc|cc|cc|cc}
\hline
       & \multicolumn{12}{c}{K} \\
\cline{2-13} 
         & \multicolumn{2}{c|}{0.01} & \multicolumn{2}{c|}{0.05} & \multicolumn{2}{c|}{0.10} & \multicolumn{2}{c|}{0.15} & \multicolumn{2}{c|}{0.20} & \multicolumn{2}{c}{0.25} \\ 
\cline{2-13} 
         & mean  & std   & mean  & std   & mean  & std   & mean  & std   & mean  & std   & mean  & std   \cr
\hline \hline
RAZOR    & 0.026 & 0.0044 & 0.063 & 0.0030 & 0.110 & 0.0017 & 0.160 & 0.0014 & 0.208 & 0.0011 & 0.258 & 0.0013 \cr 
\hline
ENRICH   & 0.008 & 0.0083 & 0.041 & 0.0204 & 0.086 & 0.0326 & 0.133 & 0.0365 & 0.176 & 0.0302 & 0.218 & 0.0281 \cr 
\hline
CIS      & 0.010 & 0.0024 & 0.049 & 0.0083 & 0.099 & 0.0125 & 0.149 & 0.0159 & 0.199 & 0.0176 & 0.248 & 0.0210 \cr 
\hline
GDIS     &   -   &   -    &   -   &   -    &   -   &   -    & 0.159 & 0.0340 &   -   &   -    &   -   &   -    \cr 
\hline
EGDIS    &   -   &   -    &   -   &   -    &   -   &   -    & 0.141 & 0.0354 &   -   &   -    &   -   &   -    \cr 
\hline
\label{tab:results-hist-balancing}
\end{tabular}
\end{table*}
For network evaluation, the "CRC-VAL-HE-7K" dataset was used, consisting of 7180 image patches derived from 50 colorectal adenocarcinoma patients, with no overlap with the patients from the "NCT-CRC-HE-100K" dataset. This dataset, serving as a validation set for models trained on the larger dataset, maintains the same image characteristics in terms of size (224x224 px) and resolution (0.5 MPP). All tissue samples were provided by the NCT biobank, ensuring consistency with the ethical and procedural details of the training dataset.

During the experiments, subsets of images of varying sizes were selected from the "NCT-CRC-HE-100K" dataset using various instance selection methods. These subsets were used to train three different deep learning networks: AlexNet, VGG16, and ResNet18. The trained networks were subsequently evaluated using the "CRC-VAL-HE-7K" dataset on a multi-class classification problem. This methodology allowed for comparing the effectiveness of different sample selection methods in terms of accuracy and F1-score, providing a robust assessment of the performance of the trained models.

First of all, the time required for different methods to select a representative subset of training instances from the entire dataset was measured, as this time needs to be added to the time required for the various architectures to be trained on the reduced dataset. Table~\ref{tab:results-histo-seltimes} shows the time in hours taken by each method to select different percentages of instances from the entire dataset. It can be observed that techniques like ENRICH take so long that the selection time alone exceeds the time required to train the architectures on the full dataset, seemingly negating the utility of applying a selection technique. However, applying selection techniques might still be reasonable if the original dataset is not very large or if multiple architectures need to be trained on the extracted subset of instances. Moreover, as can be observed from the data in Table~\ref{tab:results-histo-seltimes}, the reported times for RAZOR remain nearly identical regardless of the value of $k$. This can be explained by the fact that the majority of RAZOR's runtime is spent during the clustering phase of the instances, whereas the actual selection process has a negligible impact on the overall computational time.

\begin{table}[h]
\centering
\caption{Performance and training times (in hours) on the classification task for Different CNN Architectures when trained from scratch (FS) or fine tuned (FT) on the whole training set. }
\begin{tabular}{l|l|*{2}{c|}*{2}{c|}*{2}{c}}
\hline
       &        & \multicolumn{2}{c|}{AlexNet} & \multicolumn{2}{c|}{VGG16} & \multicolumn{2}{c}{ResNet18} \\
\cline{3-8}
\cline{3-8}
       &        & FS & FT & FS & FT & FS & FT \\
\hline \hline
\textbf{Accuracy}  &   & 0.93 & 0.92 & 0.94 & 0.96 & 0.93 & 0.91 \\ \hline
\textbf{F1-Score}  &   & 0.91 & 0.90 & 0.91 & 0.95 & 0.91 & 0.89 \\ \hline
\textbf{Training Time}  &   & 4.7 & 5.1 & 10.7 & 12.6 & 4.3 & 5.5 \\ \hline
\end{tabular}
\label{tab:cnn_performance}
\end{table}

An important aspect to evaluate in different instance selection techniques is the balance of the extracted subset with respect to the number of classes. To this aim, for each value of $k$, the percentage of samples extracted from each class was calculated, and the mean and variance of these values were computed. A lower variance indicates a more balanced reduced subset, meaning that the instance selection technique better distributes samples across the classes. The results of this analysis are presented in Table~\ref{tab:results-hist-balancing}.

\begin{table*}[h]
\centering
\caption{Accuracy, F1-score, and Training Time Results for Different Networks and Values of $k$ when training from scratch.}
\begin{tabular}{l|l|*{6}{c|}*{6}{c|}*{6}{c}}
\hline
       &        & \multicolumn{6}{c|}{AlexNet} & \multicolumn{6}{c|}{VGG16} & \multicolumn{6}{c}{ResNet18} \\
\cline{3-20}
       &        & \multicolumn{6}{c|}{K} & \multicolumn{6}{c|}{K} & \multicolumn{6}{c}{K} \\
\cline{3-20}
       &        & 0.01 & 0.05 & 0.10 & 0.15 & 0.20 & 0.25 & 0.01 & 0.05 & 0.10 & 0.15 & 0.20 & 0.25 & 0.01 & 0.05 & 0.10 & 0.15 & 0.20 & 0.25 \\
\hline \hline
RAZOR       & Acc. & 0.92 & 0.96 & 0.95 & 0.94 & 0.95 & 0.96 & 0.95 & 0.94 & 0.96 & 0.96 & 0.96 & 0.96 & 0.91 & 0.92 & 0.93 & 0.94 & 0.94 & 0.94 \\
               & F1  & 0.90 & 0.94 & 0.93 & 0.95 & 0.95 & 0.95 & 0.92 & 0.91 & 0.95 & 0.94 & 0.95 & 0.93 & 0.88 & 0.89 & 0.92 & 0.92 & 0.92 & 0.92 \\
               & Hours & 0.19 & 0.67 & 1.30 & 1.93 & 2.78 & 3.67 & 0.16 & 0.55 & 1.09 & 1.69 & 2.55 & 3.52 & 0.05 & 0.11 & 0.31 & 0.40 & 0.71 & 1.41 \\
\hline
Enrich         & Acc. & 0.68 & 0.90 & 0.93 & 0.95 & 0.96 & 0.94 & 0.61 & 0.89 & 0.95 & 0.95 & 0.92 & 0.93 & 0.58 & 0.89 & 0.93 & 0.93 & 0.92 & 0.93 \\
               & F1  & 0.59 & 0.88 & 0.93 & 0.92 & 0.95 & 0.92 & 0.54 & 0.87 & 0.93 & 0.93 & 0.88 & 0.91 & 0.51 & 0.87 & 0.91 & 0.91 & 0.89 & 0.91 \\
               & Hours & 0.17 & 0.66 & 1.28 & 1.91 & 2.72 & 3.64 & 0.15 & 0.54 & 1.08 & 1.67 & 2.53 & 3.50 & 0.03 & 0.10 & 0.29 & 0.39 & 0.71 & 1.40 \\
\hline
CIS            & Acc. & 0.91 & 0.94 & 0.93 & 0.95 & 0.95 & 0.95 & 0.91 & 0.93 & 0.94 & 0.95 & 0.95 & 0.93 & 0.88 & 0.91 & 0.93 & 0.92 & 0.93 & 0.93 \\
               & F1  & 0.88 & 0.92 & 0.92 & 0.92 & 0.93 & 0.94 & 0.89 & 0.90 & 0.92 & 0.93 & 0.93 & 0.91 & 0.84 & 0.88 & 0.91 & 0.89 & 0.91 & 0.91 \\
               & Hours & 0.18 & 0.66 & 1.29 & 1.92 & 2.74 & 3.65 & 0.16 & 0.54 & 1.08 & 1.68 & 2.53 & 3.51 & 0.03 & 0.10 & 0.30 & 0.39 & 0.71 & 1.41 \\
\hline
GDIS           & Acc. & - & - & - & 0.91 & - & - & - & - & - & 0.93 & - & - & - & - & - & 0.92 & - & - \\
               & F1  & - & - & - & 0.90 & - & - & - & - & - & 0.90 & - & - & - & - & - & 0.89 & - & - \\
               & Hours & - & - & - & 1.93 & - & - & - & - & - & 1.95 & - & - & - & - & - & 0.72 & - & - \\
\hline
EGDIS          & Acc. & - & - & - & 0.92 & - & - & - & - & - & 0.92 & - & - & - & - & - & 0.91 & - & - \\
               & F1  & - & - & - & 0.90 & - & - & - & - & - & 0.90 & - & - & - & - & - & 0.88 & - & - \\
               & Hours & - & - & - & 1.96 & - & - & - & - & - & 1.46 & - & - & - & - & - & 0.28 & - & - \\
\hline
\end{tabular}
\label{tab:results-histo-fs}
\end{table*}

The results in Table~\ref{tab:results-hist-balancing} demonstrate that RAZOR consistently delivers the most balanced and stable performance across all $k$ values, with notably low standard deviations. This highlights its effectiveness in selecting a well-balanced subset of samples across all classes. In contrast, ENRICH exhibits significantly higher variability, particularly at lower $k$ values, indicating that it struggles to maintain balance when selecting fewer samples. CIS, while generally more stable than ENRICH, shows an increasing trend in standard deviation as $k$ increases, suggesting it may encounter difficulties in maintaining balance as the sample size grows. GDIS and EGDIS, which automatically determine $k$ without allowing it to be set as a parameter, both show relatively high standard deviations, further reflecting their challenges in ensuring balanced sample selection.

In order to assess how different instance selection techniques impact the effectiveness and efficiency of CNN architectures designed for classification tasks, the performance of three CNN models (AlexNet, VGG16, and ResNet18) was evaluated. The models were trained using the entire training set, and the results are reported in Table~\ref{tab:cnn_performance}. This evaluation highlights the comparative performance of these architectures, offering insights into how well each network 
behaves when trained on complete data, thus providing a baseline for understanding the potential benefits or drawbacks of applying instance selection methods.

The performances of 
the chosen 
IS methods (RAZOR, Enrich, CIS, GDIS, and EGDIS) across the different CNN architectures (AlexNet, VGG16, and ResNet18) with varying values of 
$k$ are reported in Table~\ref{tab:results-histo-fs} and Table~\ref{tab:results-histo-ft} in terms of accuracy, F1-score, and training time. Table~\ref{tab:results-histo-fs} presents the results for the CNN architectures when trained from scratch on a classification task, using a variable portion $k$ ranging from 0.01 to 0.25 of the original training samples. Table~\ref{tab:results-histo-ft} displays the results for the same CNN architectures, pre-trained on ImageNet, and fine-tuned on a classification task with the same range of $k$ values.

\begin{table*}[ht]
\centering
\caption{Accuracy, F1-score, and Training Time Results for Different Networks and Values of $k$ on the fine-tuning task.} 
\begin{tabular}{l|l|*{6}{c|}*{6}{c|}*{6}{c}}
\hline
       &        & \multicolumn{6}{c|}{AlexNet} & \multicolumn{6}{c|}{VGG16} & \multicolumn{6}{c}{ResNet18} \\
\cline{3-20}
       &        & \multicolumn{6}{c|}{K} & \multicolumn{6}{c|}{K} & \multicolumn{6}{c}{K} \\
\cline{3-20}
         &    & 0.01 & 0.05 & 0.10 & 0.15 & 0.20 & 0.25 & 0.01 & 0.05 & 0.10 & 0.15 & 0.20 & 0.25 & 0.01 & 0.05 & 0.10 & 0.15 & 0.20 & 0.25 \\
\hline \hline
RAZOR       & Acc. & 0.95 & 0.96 & 0.96 & 0.96 & 0.96 & 0.96 & 0.92 & 0.96 & 0.95 & 0.96 & 0.96 & 0.96 & 0.90 & 0.94 & 0.94 & 0.94 & 0.94 & 0.94 \\
               & F1  & 0.94 & 0.94 & 0.94 & 0.95 & 0.93 & 0.95 & 0.89 & 0.94 & 0.94 & 0.95 & 0.95 & 0.95 & 0.85 & 0.91 & 0.92 & 0.93 & 0.92 & 0.92 \\
               & Hours & 0.21 & 0.67 & 1.36 & 1.98 & 3.13 & 4.77 & 0.17 & 0.56 & 1.15 & 1.75 & 2.34 & 3.25 & 0.03 & 0.11 & 0.27 & 0.43 & 0.61 & 1.32 \\
\hline
Enrich         & Acc. & 0.58 & 0.87 & 0.95 & 0.96 & 0.94 & 0.95 & 0.65 & 0.91 & 0.93 & 0.95 & 0.94 & 0.95 & 0.64 & 0.89 & 0.91 & 0.93 & 0.92 & 0.92 \\
               & F1  & 0.45 & 0.84 & 0.93 & 0.94 & 0.92 & 0.94 & 0.56 & 0.88 & 0.91 & 0.92 & 0.93 & 0.93 & 0.51 & 0.86 & 0.88 & 0.91 & 0.90 & 0.89 \\
               & Hours & 0.19 & 0.64 & 1.32 & 1.94 & 3.07 & 4.50 & 0.16 & 0.54 & 1.14 & 1.71 & 2.32 & 3.21 & 0.03 & 0.10 & 0.26 & 0.41 & 0.59 & 0.25 \\
\hline
CIS            & Acc. & 0.92 & 0.93 & 0.95 & 0.96 & 0.94 & 0.94 & 0.89 & 0.93 & 0.95 & 0.94 & 0.96 & 0.95 & 0.87 & 0.93 & 0.92 & 0.92 & 0.92 & 0.94 \\
               & F1  & 0.89 & 0.90 & 0.94 & 0.94 & 0.91 & 0.91 & 0.86 & 0.88 & 0.93 & 0.93 & 0.94 & 0.93 & 0.82 & 0.90 & 0.90 & 0.90 & 0.90 & 0.92 \\
               & Hours & 0.21 & 0.67 & 1.36 & 2.98 & 3.14 & 4.77 & 0.17 & 0.56 & 1.15 & 1.75 & 2.34 & 3.25 & 0.04 & 0.11 & 0.27 & 0.43 & 0.61 & 1.32 \\
\hline
GDIS           & Acc. & - & - & - & 0.93 & - & - & - & - & - & 0.93 & - & - & - & - & - & 0.92 & - & - \\
               & F1  & - & - & - & 0.91 & - & - & - & - & - & 0.91 & - & - & - & - & - & 0.90 & - & - \\
               & Hours & - & - & - & 1.98 & - & - & - & - & - & 1.56 & - & - & - & - & - & 0.87 & - & - \\
\hline
EGDIS          & Acc. & - & - & - & 0.91 & - & - & - & - & - & 0.93 & - & - & - & - & - & 0.92 & - & - \\
               & F1  & - & - & - & 0.88 & - & - & - & - & - & 0.90 & - & - & - & - & - & 0.90 & - & - \\
               & Hours & - & - & - & 1.93 & - & - & - & - & - & 1.39 & - & - & - & - & - & 0.28 & - & - \\
\hline
\end{tabular}
\label{tab:results-histo-ft}
\end{table*}

The results in Table~\ref{tab:results-histo-fs} and~\ref{tab:results-histo-ft} show that RAZOR consistently outperforms the other methods in terms of both accuracy and F1-score across all tested values of $k$. RAZOR’s ability to maintain high performance across varying architectures and sample sizes underscores its effectiveness in selecting high-quality, representative samples that enhance the training process from scratch and continue to benefit the fine-tuning phase. The slight variations observed in RAZOR's results further suggest its robustness and adaptability across different conditions, making it a reliable choice for CNN training. In contrast, Enrich, while performing well, particularly with larger sample sizes (higher values of $k$), exhibits greater variability in its results. This inconsistency may be attributed to its sensitivity to the initial sample selection, which could introduce biases or redundancies, affecting the overall performance. Although Enrich improves during the fine-tuning phase, its variability indicates that it may not be as reliable as RAZOR across all scenarios. CIS demonstrates competitive performance, particularly in certain conditions, but generally falls short of RAZOR.
\begin{table*}[h!]
\centering
\caption{Performance scores for the UNet binary mask segmentation task.}
\begin{tabular}{lcccccc|cccccc}
\hline
 & \multicolumn{6}{c|}{RAZOR} & \multicolumn{6}{c}{Enrich} \\
 \hline
 & \multicolumn{6}{c|}{K} & \multicolumn{6}{c}{K} \\
 \hline
 & 0.01 & 0.05 & 0.10 & 0.15 & 0.20 & 0.25 & 0.01 & 0.05 & 0.10 & 0.15 & 0.20 & 0.25 \\
\hline
\hline
F1Score     & 0.77 & 0.80 & 0.83 & 0.84 & 0.84 & 0.84 & 0.73 & 0.77 & 0.78 & 0.79 & 0.80 & 0.80 \\
\hline
IoU         & 0.72 & 0.74 & 0.77 & 0.77 & 0.78 & 0.77 & 0.69 & 0.71 & 0.72 & 0.73 & 0.74 & 0.74 \\
\hline
Dice        & 0.77 & 0.80 & 0.83 & 0.84 & 0.84 & 0.84 & 0.73 & 0.77 & 0.78 & 0.80 & 0.80 & 0.79 \\
\hline
Overlap Rate & 0.78 & 0.81 & 0.83 & 0.84 & 0.85 & 0.84 & 0.74 & 0.78 & 0.79 & 0.80 & 0.81 & 0.80 \\
\hline
Hours       & 0.13 & 0.83 & 1.68 & 2.89 & 4.07 & 5.30 & 0.15 & 0.80 & 1.62 & 2.76 & 3.94 & 5.10 \\
\hline
\end{tabular}
\label{tab:results_ubm}
\end{table*}

Its strategy of selecting central instances within clusters may help focus on representative samples, especially during fine-tuning. However, the underperformance of CIS compared to RAZOR suggests that it may not capture the diversity necessary for robust training, leading to lower effectiveness in some cases. GDIS and EGDIS, which are density-based selection methods, show limited results and generally underperform compared to both RAZOR and CIS. Their lower overall effectiveness implies that their approach may not be as suitable for scenarios requiring extensive fine-tuning, where diversity in the training samples is crucial. Moreover, GDIS and EGDIS do not allow for explicit specification of the portion of samples $k$ to be selected; instead, the selected portion results directly from the method itself. In the experiments, both GDIS and EGDIS produced highly variable selection portions across the nine classes in the dataset, ranging from a minimum of $0.11$ to a maximum of $0.19$ for GDIS, and from $0.08$ to $0.17$ for EGDIS. This variability resulted in a disproportionality between the most and least represented classes, which was approximately double for both techniques. Given that both methods generally select an average of 0.15 of the entire dataset, their performance was only reported for $k=0.15$. This characteristic significantly limits the flexibility of these methods.

Additionally, the statistical significance of the obtained results was assessed through the Wilcoxon test, which further corroborates the superiority of RAZOR over the other methods. The Wilcoxon test results show significant differences in performance between RAZOR and the compared methods—Enrich, CIS, GDIS, and EGDIS—with p-values of 0.0054, 0.0215, 0.0353, and 0.0236, respectively, for accuracy. Similarly, for the F1-score, the p-values were 0.0058, 0.0193, 0.0412, and 0.0321, respectively. These values indicate that the differences in performance are not due to random chance, thereby statistically validating RAZOR's superior performance across all tested metrics.

\subsubsection{MRI Volumes}
Brain MRI images were selected to validate segmentation and multi-pixel classification tasks, since the segmentation of brain MRIs presents several challenges, including intensity inhomogeneity, complex anatomical structures, and the presence of noise and artifacts. Furthermore, the high dimensionality and significant data volumes associated with MRI datasets require substantial computational resources for processing and analysis.

The MRI dataset used for the experiments consists of 142 anatomical volumes (from 76 unique subjects) obtained using an MP2RAGE sequence at an isotropic resolution of 0.63 $mm^3$, utilizing a 7 Tesla MRI scanner with a 32-channel head coil. All volumes were collected as reconstructed DICOM images at the Imaging Centre of Excellence (ICE) at the Queen Elizabeth University Hospital in Glasgow, UK. For each volume, corresponding INV1, INV2, and UNI\_Images scans were released. Additionally, automatic segmentations obtained using FreeSurfer~\cite{fischl2012freesurfer} and a custom tool~\cite{fracasso2016lines}, as well as manual segmentations used for a behavioral survey, were made available. No preprocessing steps were applied to the data.

The anatomical data follows the BIDS format. Along with the publicly available data, a 6-class segmentation mask is provided. The segmented classes are: background, gray matter, basal ganglia, white matter, ventricles, cerebellum, and brainstem.

The ground truth was obtained from the FreeSurfer recon-all procedure, merging classes such that only those used in the MICCAI MRBrainS13 and MRBrainS18 challenges were retained, except for the less numerous classes ("White matter lesions", "Infarction", and "Other") which were not directly obtainable.

\begin{table*}[h!]
\centering
\caption{Performance scores for the UNet multi-pixel classification task}
\begin{tabular}{lcccccc|cccccc}
\hline
 & \multicolumn{6}{c|}{RAZOR} & \multicolumn{6}{c}{Enrich} \\
 \hline
 & \multicolumn{6}{c|}{K} & \multicolumn{6}{c}{K} \\
 \hline
 & 0.01 & 0.05 & 0.10 & 0.15 & 0.20 & 0.25 & 0.01 & 0.05 & 0.10 & 0.15 & 0.20 & 0.25 \\
\hline
\hline
F1Score     & 0.73 & 0.80 & 0.85 & 0.85 & 0.85 & 0.86 & 0.65 & 0.69 & 0.75 & 0.80 & 0.78 & 0.81 \\
\hline
IoU         & 0.69 & 0.75 & 0.81 & 0.80 & 0.80 & 0.81 & 0.62 & 0.66 & 0.72 & 0.75 & 0.73 & 0.77 \\
\hline
Dice        & 0.73 & 0.80 & 0.85 & 0.85 & 0.85 & 0.86 & 0.65 & 0.69 & 0.75 & 0.80 & 0.78 & 0.81 \\
\hline
Overlap Rate & 0.84 & 0.86 & 0.88 & 0.88 & 0.89 & 0.86 & 0.73 & 0.83 & 0.84 & 0.85 & 0.86 & 0.85 \\
\hline
Hours       & 0.15 & 0.72 & 2.69 & 4.94 & 5.31 & 7.12 & 0.16 & 0.77 & 2.77 & 5.06 & 5.74 & 7.33 \\
\hline
\end{tabular}
\label{tab:results_upc}
\end{table*}

The first experiment focuses on the binary segmentation of brain MRI images. The dataset consists of 76 MR volumes, which are split into two disjoint sets based on patients: the first 66 volumes are used for training, while the remaining 10 are reserved for testing. Each MR volume is divided into individual slices, and each slice is further partitioned into non-overlapping patches of size 64x64 pixels. Patches containing more than 75\% background are excluded, resulting in a total of 343,168 training patches and 49,725 testing patches. A UNET network is trained from scratch to perform binary segmentation on these patches. The performance metrics considered are F1-score, Intersection over Union (IoU), Dice, and Overlap Rate.

The second experiment involves multi-class pixel classification in MRI volumes, targeting the six classes provided by the dataset. The setup mirrors the first experiment, but in this case, the UNET network is configured to perform a regression task across six distinct classes. Performance is evaluated using metrics such as F1-score, Intersection over Union (IoU), Dice, and Overlap Rate. These metrics are computed individually for each class and then averaged to provide an overall performance measure across all classes.

For the binary segmentation task, the UNET network required 24.36 hours of training and achieved the following performance on the testing set: an F1-score of 0.85, an IoU of 0.78, a Dice coefficient of 0.85, and an Overlap Rate of 0.86. In the multi-class pixel classification task, the UNET network required 50.18 hours of training. On the testing set, it achieved an F1-score of 0.85, IoU of 0.80, Dice coefficient of 0.85, and an Overlap Rate of 0.89. 

In these two experiments, the individual patches do not have labels themselves, but the pixels within them carry labels. Therefore, the only two instance selection techniques that do not require labels on the images are RAZOR and ENRICH, which are, consequently, the methods considered for comparisons. For this task, RAZOR requires 8.25 hours (one core) and 4.23 hours (four cores) independent from the percentage $k$ of selected data. Instead, for ENRICH, time usages vary greatly with $k$: 7.20, 30.33, 75.90, 115.56, 165.36, 195.57 hours are required for values of $k$ equals, respectively, to 0.01, 0.05, 0.10, 0.15, 0.20, 0.25, 


The performance of the UNet network in segmenting MR images into binary masks using two different sample selection methods, RAZOR and Enrich, with progressive sample percentages ($K$), are reported in Table~\ref{tab:results_ubm}.

The results highlight that RAZOR exhibits higher performance than Enrich across all metrics. This suggests that RAZOR consistently selects more representative and relevant training samples, which leads to better segmentation outcomes. Moreover, RAZOR’s performance stabilizes after $k=0.10$, suggesting that it is effective even with smaller training sample sizes. Enrich, on the other hand, shows greater variability, particularly at lower $k$ values, indicating that it may not be as robust in maintaining segmentation accuracy with fewer training samples. Both methods show diminishing improvements as $k$ increases suggesting that increasing the sample size beyond a certain point provides little added benefit, especially for RAZOR, which maintains high performance even with smaller sample sizes.

Furthermore, the statistical analysis confirms the superiority of RAZOR, with the Wilcoxon test showing that the differences between the two methods are statistically significant. The p-values obtained by comparing performance metrics between the two methods are all below 0.05, demonstrating that the observed differences are not due to chance but are statistically significant. Specifically, for the F1-score, the p-value is 0.0054, for IoU it is 0.0071, for Dice it is 0.0065, and for Overlap Rate it is 0.0098.

The results of the second experiment on multi-class pixel classification are presented in Table~\ref{tab:results_upc}. This table illustrates the performance of the UNet network, comparing the RAZOR and Enrich methods across different values of $k$.

The results show that RAZOR demonstrates a clear advantage over Enrich at very low values of $k$. This suggests that RAZOR is more effective at selecting informative samples, even when the dataset is significantly reduced, which is particularly important in resource-constrained environments. As $k$ increases to 0.25, the performance gap between RAZOR and Enrich narrows, especially in metrics like IoU and Dice. This indicates that while RAZOR maintains consistent superiority across most $k$ values, Enrich's performance becomes more competitive as the sample size increases. However, despite this improvement, the longer execution times for Enrich make the method less practical and, in some cases, even impractical for very large datasets. 
Therefore, while Enrich shows potential with larger sample sizes, its efficiency is compromised by its high computational cost.
Furthermore, the Wilcoxon test results confirm the statistical superiority of RAZOR over Enrich. The p-values obtained by comparing performance metrics between the two methods are all below 0.05, demonstrating that the observed differences are not due to chance but are statistically significant. Specifically, for the F1-score, the p-value is 0.0058, for IoU it is 0.0071, for Dice it is 0.0065, and for Overlap Rate it is 0.0098.

\section*{Conclusions}

In this work, we have proposed RAZOR a novel instance selection method, which is based on an iterative split-and-merge clustering algorithm that leverages the concept of entropy. RAZOR has been specifically designed to be robust, efficient, and scalable, with the primary goal of facilitating its application to large-scale datasets. Unlike many techniques in the literature, RAZOR is capable of operating in both supervised and unsupervised settings. In the absence of labeled data, the clustering process groups different samples into implicit classes based on their affinity, from which a representative subset of instances is selected.

The experimental results demonstrate the superiority of RAZOR over recent state-of-the-art techniques, both in terms of effectiveness and efficiency. 
All this evidence together suggests that RAZOR might be considered a general-purpose algorithm suitable for different tasks on many dataset types and confirm its potential as a powerful tool for instance selection, particularly in scenarios involving large and complex datasets.

\section*{Acknowledgments}
We acknowledge financial support from the project PNR MUR project PE0000013-FAIR


\bibliographystyle{IEEEtran}
\bibliography{RAZOR}
\section{Biography Section}
\begin{IEEEbiography}[{\includegraphics[width=1in,height=1.25in,clip,keepaspectratio]{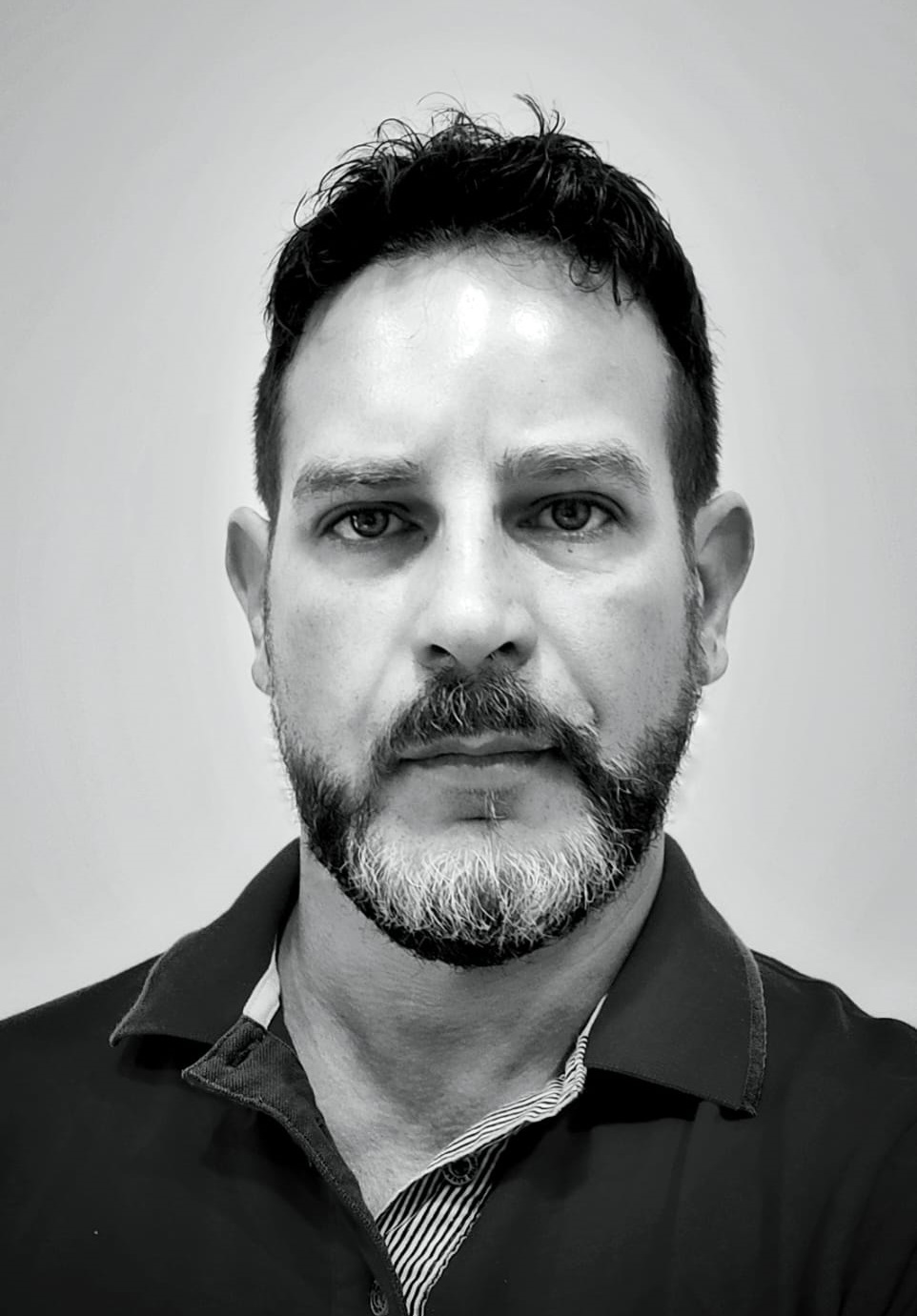}}]{Daniel Riccio} received the Laurea degree (cum laude) and the Ph.D. degree in computer sciences from the University of Salerno, Salerno, Italy, in 2002 and 2006, respectively. He is currently an Associate Professor with the University of Naples Federico II. He is also an Associate Researcher with the National Research Council of Italy. His research interests include biometrics, medical imaging, image processing and indexing, and image and video analytics. He is an IEEE Member and a member of the Italian Association of Computer Vision, Pattern Recognition, and Machine Learning (CVPL).
\end{IEEEbiography}

\begin{IEEEbiography}[{\includegraphics[width=1in,height=1.25in,clip,keepaspectratio]{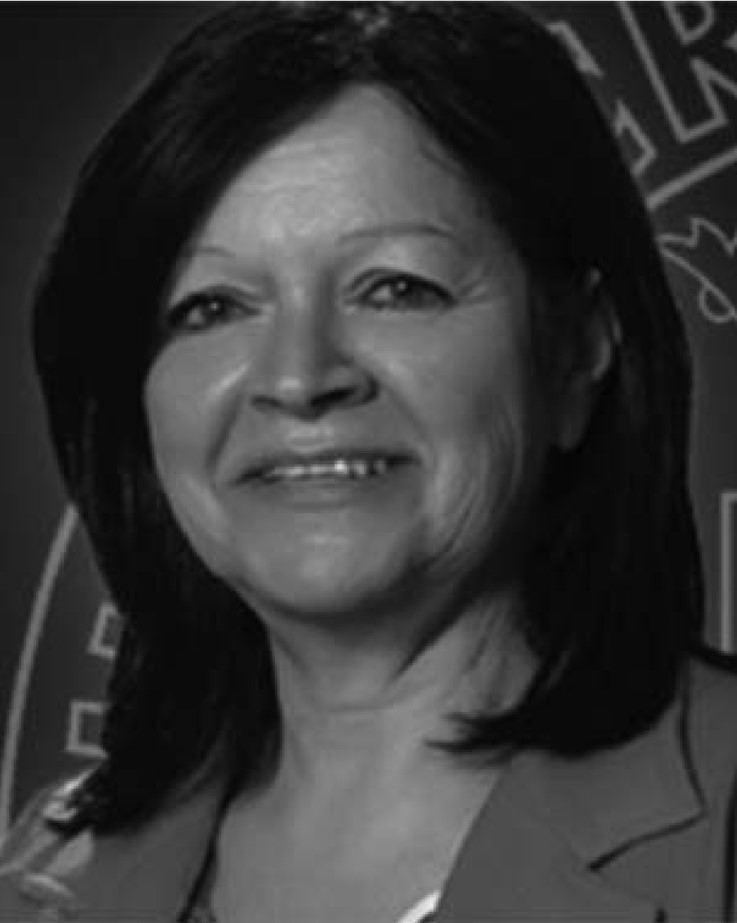}}]{Genoveffa Tortora} (Senior Member, IEEE) was the Dean of the Faculty of Mathematical, Natural, and Physical Sciences, University of Salerno, Fisciano, Italy, from 2000 to 2008. Since 1990, he has been a Full Professor of computer science. She is the Scientific Director of the Laboratory on context-aware intelligent systems, Department of Computer Science. She is the author and coauthor of more than 290 papers published in scientific journals or proceedings of refereed conferences and is Co-Editor of three books. Her research interests include software engineering, visual languages and human–machine interaction, image processing and biometric systems, Big Data, data warehouses, data mining, and geographic information systems.
Ms. Tortora is currently an Editorial Board Member of high-quality international journals, Steering Committee Member of the International Working Conference on Advanced Visual Interfaces held in cooperation with ACM, and Program Chair and Program Committee Member of several relevant international conferences. She is an ACM Member, European Association of Theoretical Computer Science Member, International Association of Pattern Recognition Reviewer for several international scientific journals, Evaluator of research projects for Italian Ministries, Regions, Universities, and European Commission Member of the Board of Examiners for several Researcher, Associate Professor, and Full Professor positions both at a national and an international level.
\end{IEEEbiography}

\begin{IEEEbiography}[{\includegraphics[width=1in,height=1.25in,clip,keepaspectratio]{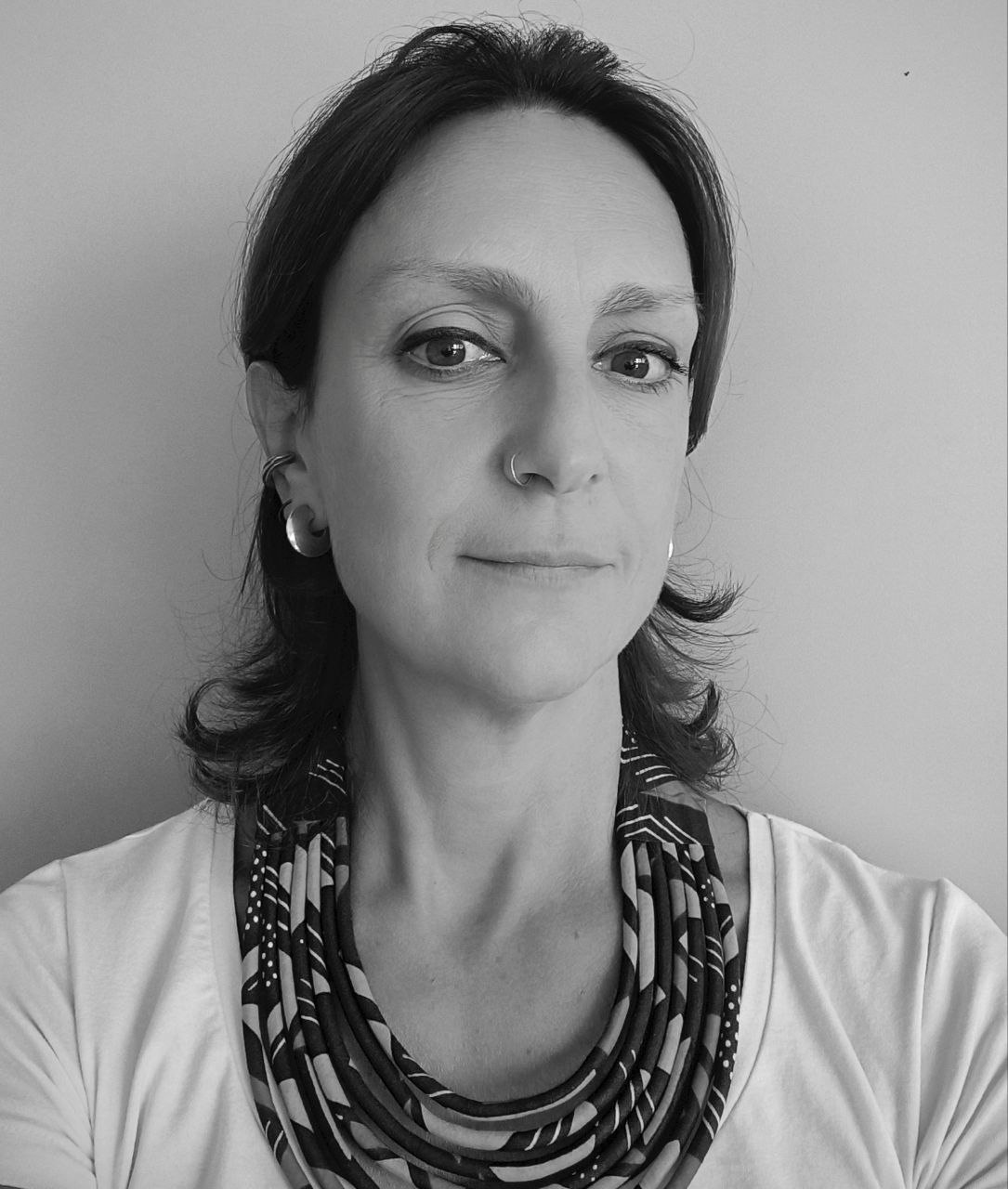}}]{Mara Sangiovanni} received the Master’s degree in Computer Science (cum laude) in 2009 at Federico II University of Naples. In 2014, she received a Ph.D. in Computational Biology and Bioinformatics. She worked on applying machine learning methodologies to the classification of genomics and transcriptomics data produced by Next Generation Sequencing techniques. 
From July 2020 to October 2023, she was a researcher at the ICAR-CNR in the Artificial Intelligence in Image and Signal Analysis Lab. Now, she is a Researcher at Federico II University of Naples. Her research interests include Deep Learning models for segmenting and classifying Medical images and Machine Learning applied to Omics Data. 
\end{IEEEbiography}
\vspace{300pt}

\end{document}